\documentclass[letterpaper,journal]{IEEEtran}
\usepackage{amsmath,amsfonts}
\usepackage{algorithmic}
\usepackage{algorithm}
\usepackage{array}
\usepackage[caption=false,font=normalsize,labelfont=sf,textfont=sf]{subfig}
\usepackage{textcomp}
\usepackage{stfloats}
\usepackage{url}
\usepackage{verbatim}
\usepackage{graphicx}
\usepackage[utf8]{inputenc}
\usepackage{graphicx}
\usepackage{booktabs}
\usepackage{multirow}
\usepackage{siunitx}
\usepackage{cite}
\usepackage{multirow}
\usepackage{graphicx}
\usepackage{caption}
\usepackage{booktabs}
\usepackage{rotating}
\usepackage{array}
\usepackage{siunitx}
\usepackage{float}
\usepackage{overpic}
\usepackage{pgfplots}
\usepackage{siunitx}
\usepackage{booktabs} 
\usepackage{multirow}
\usepackage{graphicx}
\usepackage{caption}
\usepackage{color}

\usepackage{graphicx}
\usepackage{amsopn,amssymb}
\usepackage{amsthm}
\usepackage{cite}
\usepackage{makecell}
\usepackage{multirow}
\usepackage{multicol}
\usepackage{color}
\usepackage{url}
\usepackage{soul}
\usepackage{enumitem}
\usepackage{wasysym}
\usepackage{changepage}
\usepackage{bm}
\usepackage{hyperref}
\usepackage{booktabs}
\usepackage{pifont}
\usepackage{algorithm}
\usepackage{algorithmic}
\usepackage{tikz}
\usepackage{pgfplots}
\pgfplotsset{compat=1.18}
\usepgfplotslibrary{groupplots}
\definecolor{hierBlue}{RGB}{37,99,169}
\definecolor{hierTeal}{RGB}{0,132,122}
\definecolor{hierOrange}{RGB}{218,121,35}
\definecolor{hierGray}{RGB}{105,112,120}

\def\0{{\mathbf 0}}
\def\1{{\mathbf 1}}

\usepackage{rotating}
\usepackage{array}
\usepackage{siunitx}
\usepackage{float}
\usepackage{overpic}
\usepackage{pgfplots}
\usepackage{hyperref}
\hypersetup{hidelinks}
\begin{document}

\title{HierGF: Hierarchical Gaussian Fields via Geometry–perception Message Passing for Sparse-view 3D Reconstruction}

\author{Bi'an Du, Zhimin Zhang, Daizong Liu, Baoquan Chen,~\IEEEmembership{Fellow,~IEEE}, \\and Wei Hu,~\IEEEmembership{Senior~Member,~IEEE}
\thanks{
B. Du, Z. Zhang and W. Hu are with Wangxuan Institute of Computer Technology, Peking University, No. 128, Zhongguancun North Street, Beijing, China (e-mail: pkudba@stu.pku.edu.cn, 
zm\_zhang@stu.pku.edu.cn, forhuwei@pku.edu.cn).

D. Liu is with the Institute for Math \& AI, Wuhan University, Wuhan, 430072, China. (e-mail:daizongliu@whu.edu.cn).

B. Chen is with the School of Intelligence Science and Technology, Peking
University, Beijing 100871, China. E-mail: baoquan@pku.edu.cn.

Corresponding author: Wei Hu (forhuwei@pku.edu.cn).
\protect\\ \hspace*{0.5em} 
This work was supported by the National Natural Science Foundation of China (Project Number 62595774) and the Beijing Major Science and Technology Project under Contract no. Z251100008125009.
}
}

\markboth{IEEE Transactions on Multimedia}%
{Du \MakeLowercase{\textit{et al.}}: Hierarchical Gaussian Fields}


\maketitle

\begin{abstract}

Sparse view 3D reconstruction is an important and common scenario in multimedia applications, such as augmented reality/virtual reality (AR/VR) content creation, cultural heritage digitization, and certain robotic applications, where only a limited number of randomly captured views may be available. However, sparse views contain only limited 3D information, posing two major challenges: 1) too few images are available for matching, making it difficult to build multi-view consistency; 2) insufficient view coverage leads to a lack of information in under-sampled regions, resulting in missing parts of object structure. Existing methods mostly still rely on limited reprojection errors and regularization terms, which are prone to overfitting to a single view and inconsistent appearances across views. In geometrically under-sampled regions, they often rely on heuristic density control, lacking reliable guidance and often resulting in blurring and structural holes. To address these issues, this paper proposes Hierarchical Gaussian Fields (HierGF), which revisits sparse-view reconstruction from a hierarchical geometry–perception perspective and converts limited observations into reliable self-generated supervision beyond fixed priors and heuristic density control.
In particular, we transform coarse 3D geometric information and additional 2D generative priors into structured pseudo-supervision through a two-stage geometry-perception backbone network, thereby enhancing multi-view consistency with very few input views. In addition, we introduce a learnable confidence network to guide gradients toward cross-view consistent content, and a geometrically consistent densification module to improve the reconstruction of multi-view alignment and under-sampled regions. Experiments on three sparse-view reconstruction benchmarks demonstrate that HierGF improves reconstruction quality and multi-view consistency under limited observations. On the MipNeRF360 \texttt{kitchen} scene under the 4-view setting, HierGF completes per-scene reconstruction in 632~s and post-optimization novel-view rendering in 70~s on a single NVIDIA A100 GPU, yielding lower measured runtime than the compared optimization-based baselines. Additional quantitative comparisons on LLFF and DTU further validate the method under commonly used sparse-view protocols.

\end{abstract}


\section{Introduction}
\label{sec:intro}

\begin{figure}
\centering
\IfFileExists{second_round_review_response/figures/fig1_motivation_revised.pdf}{%
    \includegraphics[width=\columnwidth]{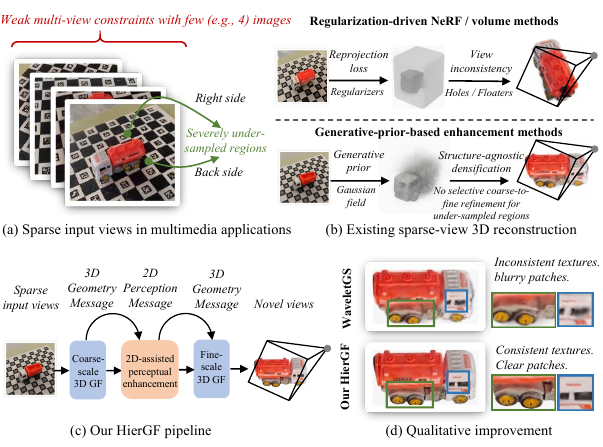}%
}{%
    \rule{0pt}{0.43\columnwidth}%
}
\caption{Motivation of HierGF. (a) Sparse inputs in multimedia scenes provide weak multi-view constraints and leave large regions under-sampled. (b) Existing sparse-view methods rely on fixed regularizers or heuristic generative priors, causing holes and inconsistent textures. (c–d) Our hierarchical pipeline with 2D-assisted perceptual supervision turns few views into reliable pseudo-supervision and produces sharper, more consistent reconstructions than recent Gaussian-based baselines.}
\vspace{-4mm}
\label{fig:teaser}
\end{figure}

Reconstructing and rendering 3D objects from 2D images is a fundamental problem in computer vision\cite{deng2022depthsupervised,roessle2022dense,du2026part,shi2024zerorf,song2026multi,li2026interaction,gao2026pre,li2025geometry,yang2024gaussianobject,gao2025hieroctfusion}. Essentially, the process involves inferring complete scene geometry, reflectivity, and appearance from a small number of inherently incomplete and projected images, supporting a wide range of applications. 
In augmented reality, this enables convincing interactions between virtual assets and the physical world \cite{azuma1997survey,cao2026electrothermally}. In robotics, this provides actionable spatial understanding for grasping, manipulation, and navigation \cite{cadena2017past,newcombe2011kinectfusion}. In the creative industry, this accelerates asset creation for games and visual effects \cite{muller2022instant,kerbl20233d,du2024generative,lai2025enhancing,du2024multi,zhang2025behave}. The ultimate goal of this problem is to produce detailed and coherent 3D models that can be inspected and re-rendered from arbitrary viewpoints. However, many 3D reconstruction methods \cite{li2018hole,mildenhall2020nerf,stankiewicz2018free,wu2016fast} still rely on dense and meticulously calibrated multi-view image acquisition. This dependence limits their application in mobile, outdoor, or time-constrained scenarios, where the number of available viewpoints is limited. Therefore, specialized methods are needed to address the 3D reconstruction problem in sparse scenes.

Early sparse-view reconstruction mostly relied on light-field, volumetric methods, and Structure from Motion (SfM) / Multi-view Stereo (MVS) pipelines. These methods require dense captures, which degrade under complex lighting, specularities, and occlusions \cite{kutulakos2000spacecarving,wilburn2005cameraarray,du2021self,seitz1999voxelcoloring,chen2022deep,seitz1996viewmorphing,buehler2001ulr}. Neural Radiance Fields (NeRF) \cite{mildenhall2020nerf} opened a new direction. Few-shot variants such as DS-NeRF \cite{deng2021dsnerf}, RegNeRF \cite{niemeyer2022regnerf}, FreeNeRF \cite{yang2023freenerf} and SparseNeRF \cite{wang2023sparsenerf} introduce depth supervision, unseen-view regularization, frequency control and depth-ranking constraints to improve stability under sparse inputs, respectively. Nevertheless, they are still volumetric and remain multi-view inconsistent in severely under-constrained regions. 
More recently, explicit primitive-based approaches such as 3D Gaussian Splatting (3DGS) \cite{kerbl20233d} have been adapted to sparse-view settings, where only a few images are available. GaussianObject \cite{yang2024gaussianobject} exploits visual-hull priors and diffusion-based Gaussian repair to recover high-quality scenes or objects from very limited views, but still relies on shape priors and structure-agnostic densification, as illustrated in Fig.~\ref{fig:teaser}(b).
This leaves open how to turn very limited observations into reliable enhanced views with strong multi-view consistency and how to optimize Gaussian primitives in a structure-aware way in under-sampled regions.

\IEEEpubidadjcol

To this end, this paper revisits sparse-view 3D reconstruction with Hierarchical Gaussian Fields (HierGF) via geometry–perception message passing. Through 2D-assisted perception enhancement and the coarse-to-fine pipeline, we transform limited observational data into reliable self-generated supervised information and apply geometry-consistent density control, thus solving the two challenges mentioned above.
Specifically, HierGF adopts a two-field coarse-to-fine framework for sparse-view 3D Gaussian reconstruction. The coarse field first provides stable geometric and color initialization. A separately optimized fine field inherits only spatial means and RGB initialization from the coarse field, while opacity, anisotropic scale, rotation, and higher-order appearance attributes are re-estimated using enhanced pseudo-view supervision and geometry-consistent densification. This selective transfer retains useful coarse-stage structure without constraining fine-scale refinement by the complete state of an early reconstruction.

Furthermore, sparse inputs can lead to large under-sampled regions with weak image gradients, where naive density control tends to produce holes, floaters, and broken thin structures. In our framework, the optimization of the fine-scale Gaussian field not only utilizes filtered enhanced views to extend supervision beyond the original observations, but also employs a geometry-consistent densification strategy to improve the reconstruction of under-sampled regions.
Specifically, new Gaussian primitives are inserted along estimated surface directions rather than at arbitrary locations, and a normal-weighted interpolation scheme propagates structure from well-constrained areas into nearby under-constrained ones. 
This allows HierGF to refine geometry and appearance in under-sampled regions while maintaining surface continuity and recovering fine details, as illustrated in Fig.~\ref{fig:teaser}.

Overall, our contributions are summarized as follows: 
\begin{itemize}
    \item We focus on sparse-view 3D reconstruction in multimedia scenarios and revisit this problem from a hierarchical geometry–perception perspective, aiming to convert limited and uneven observations into reliable self-generated supervision beyond reprojection-only objectives and heuristic density control.
    \item We propose a two-field coarse-to-fine refinement design for sparse-view 3D Gaussian reconstruction. Its defining mechanism is selective parameter inheritance from a geometrically stable coarse field to a separately optimized fine field: only spatial means and RGB initialization are transferred, while enhanced pseudo-supervision and geometry-consistent densification refine appearance and complete under-sampled regions. This design enables reliable visual guidance and local structural completion to operate on a flexible fine-scale representation rather than being constrained by the full state of an early coarse reconstruction.
    \item We provide quantitative and qualitative evaluations on three primary sparse-view benchmarks, together with additional quantitative comparisons on LLFF and DTU and a controlled efficiency analysis. Under the evaluated MipNeRF360 \texttt{kitchen} 4-view setting, HierGF completes per-scene reconstruction in 632~s on a single NVIDIA A100 GPU, corresponding to reconstruction speedups of \(2.07\times\), \(11.12\times\), and \(2.85\times\) relative to ZeroRF, GaussianObject, and WaveletGS, respectively. It also achieves the lowest measured post-optimization rendering time among these methods.
\end{itemize}

\section{Related Works}
\label{sec:related}

\begin{table*}[t]
\centering
\caption{Direct comparison with diffusion-based sparse-view reconstruction methods.}
\label{tab:r1_c1_direct_comparison}
\small
\setlength{\tabcolsep}{4pt}
\setlength{\extrarowheight}{2pt}
\renewcommand{\arraystretch}{1.5}
\begin{tabular}{@{}
>{\raggedright\arraybackslash}p{0.12\textwidth}
>{\raggedright\arraybackslash}p{0.16\textwidth}
>{\raggedright\arraybackslash}p{0.17\textwidth}
>{\raggedright\arraybackslash}p{0.18\textwidth}
>{\raggedright\arraybackslash}p{0.28\textwidth}
@{}}
\hline
\textbf{Method} &
\textbf{Representation path} &
\textbf{Diffusion use} &
\textbf{Reliability / sparse-region handling} &
\textbf{Main distinction from HierGF} \\
\hline
3DGS-Enhancer~\cite{liu2024gsenhancer} &
Fine-tunes an initial 3DGS &
Video diffusion restores rendered novel views &
Confidence scores guide enhanced-view fine-tuning &
HierGF uses a separate fine field, selective inheritance, and explicit sparse-region completion. \\
\hline
IPSM-Gaussian~\cite{wang2024diffusionpriors} &
Optimizes 3DGS with diffusion guidance &
Inline priors rectify sparse-view score matching &
Uses depth/geometric regularization in the IPSM objective &
HierGF does not rectify SDS; it refines a selectively initialized fine field with enhanced pseudo-\allowbreak supervision and GCD. \\
\hline
HierGF &
Coarse field \(\rightarrow\) selectively initialized fine field &
Scene-adapted 2D prior enhances coarse-field pseudo-views &
Confidence weights filter enhanced losses; GCD allocates Gaussians in sparse regions &
Two-field refinement with selective inheritance and coupled perceptual/\allowbreak geometric refinement. \\
\hline
\end{tabular}
\end{table*}

\begin{figure*}
  \centering
  \IfFileExists{second_round_review_response/figures/framework_modified.pdf}{%
      \includegraphics[width=0.93\textwidth]{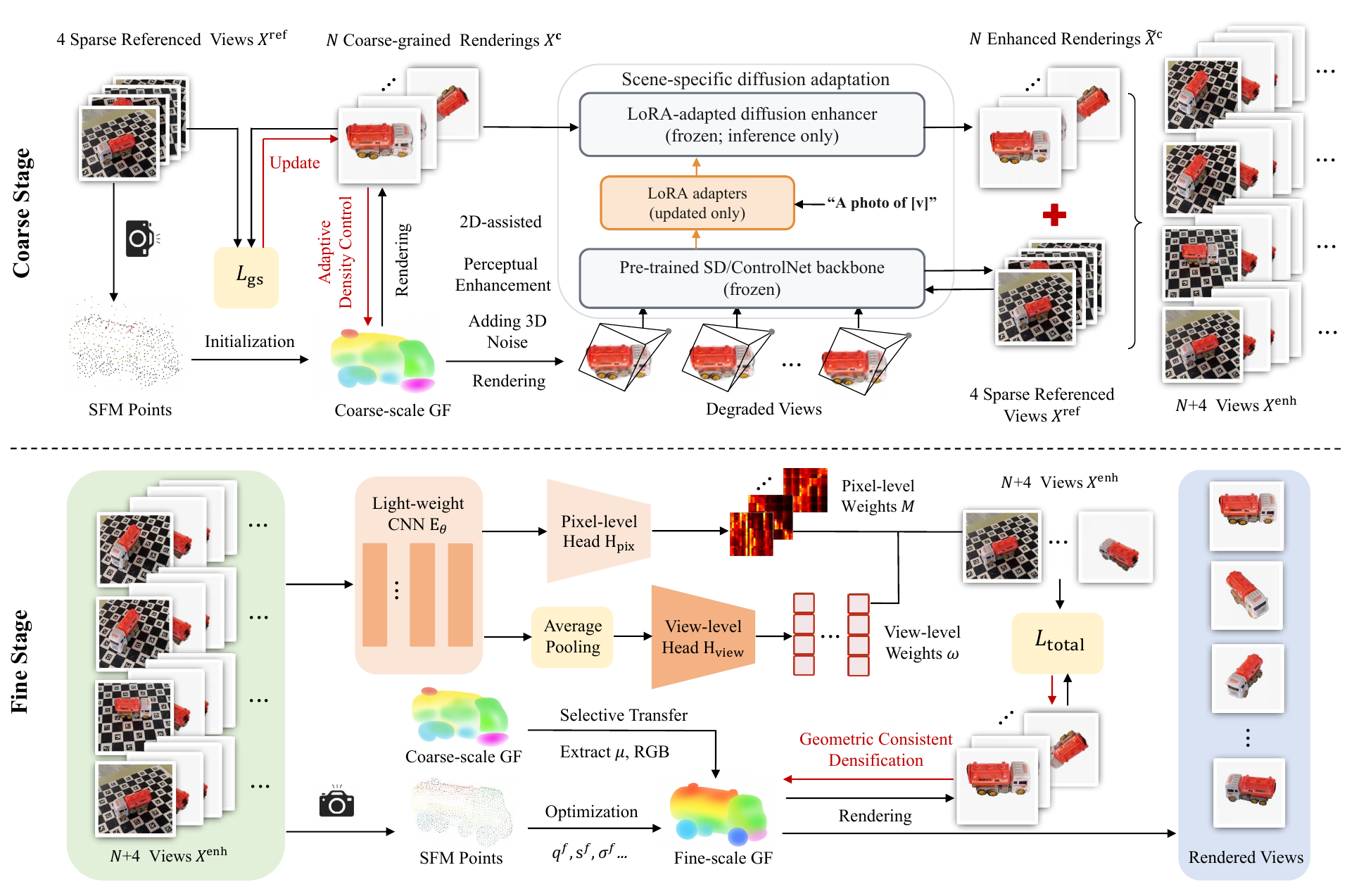}%
  }{%
      \rule{0pt}{0.38\textwidth}%
  }
  \caption{Overview of the proposed HierGF framework with explicit selective coarse-to-fine transfer and clarified diffusion adaptation. The optimized coarse Gaussian field transfers only spatial means and RGB initialization to the separately optimized fine field, whereas opacity, anisotropic scale, rotation, and higher-order appearance attributes are reinitialized. Only the LoRA adapters are updated during scene-specific diffusion adaptation; the pre-trained SD/ControlNet backbone remains frozen, and the adapted enhancer is frozen during reconstruction.}
  
  \label{fig:architecture}
  \vspace{-3mm}
\end{figure*}

We group related work into three themes: 1) optimization-based per-scene sparse-view reconstruction, where our method is situated; 2) feed-forward reconstruction and 3D priors, which use a single network across multi-scenes and are reviewed mainly for conceptual context rather than as direct baselines; and 3) reliability-aware and structural supervision, which motivate our confidence-guided supervision and geometry-consistent densification.
\subsection{Optimization-based Per-scene Sparse-view Reconstruction}
Optimization-based per-scene methods explicitly fit a 3D model for each scene. Classical space carving, voxel coloring and SfM/MVS require dense and well-calibrated views, and quickly degrade when only a few views are available \cite{kutulakos2000spacecarving,seitz1999voxelcoloring,wilburn2005cameraarray,seitz1996viewmorphing,buehler2001ulr}. NeRF \cite{mildenhall2020nerf} and its sparse-view variants \cite{deng2021dsnerf,niemeyer2022regnerf,yang2023freenerf,wang2023sparsenerf} introduce depth and unseen-view regularization to stabilize per-scene optimization under few-shot inputs, but often depend on external predictors and still struggle in under-constrained regions \cite{10814653,10500313,younis2025sparse}. 3D Gaussian Splatting (3DGS) \cite{kerbl20233d} has been adapted to the same setting. GaussianObject \cite{yang2024gaussianobject} exploits visual hulls with diffusion-based repair, while Pixel-GS, S2Gaussian and DropoutGS \cite{zhang2024pixel,wan2025s2gaussian,xu2025dropoutgs} refine density control and capacity to alleviate overfitting and improve coverage. These pipelines still rely on fixed priors and heuristic densification and ignore the reliability of synthesized supervision, which motivates our hierarchical design with 2D-assisted perceptual enhancement and geometry-consistent densification.

Recent diffusion-based sparse-view 3DGS methods improve reconstruction by introducing additional visual guidance. 3DGS-Enhancer~\cite{liu2024gsenhancer} restores view-consistent novel-view features using a video diffusion prior and confidence-aware enhanced-view supervision, while IPSM-Gaussian~\cite{wang2024diffusionpriors} rectifies score-distillation-based diffusion guidance through inline visual priors. HierGF is complementary to these approaches. Its distinction lies in a two-field coarse-to-fine refinement process: a coarse field provides a stable geometry-supported initialization and pseudo-view source, while a separately optimized fine field receives selectively inherited parameters, confidence-weighted enhanced supervision, and geometry-consistent sparse-region completion. This design focuses on how an explicit Gaussian representation is re-parameterized and refined after auxiliary visual evidence becomes available, rather than treating diffusion-enhanced views, confidence-aware supervision, or geometry-guided density control as independent modules.

Table~\ref{tab:r1_c1_direct_comparison} summarizes the representation path, diffusion usage, treatment of unreliable enhanced information, and sparse-region handling of closely related diffusion-based sparse-view reconstruction methods, clarifying that HierGF differs by combining a two-field coarse-to-fine refinement strategy with selective parameter inheritance and geometry-consistent densification.

\subsection{Feed-forward Sparse-view Reconstruction and 3D Priors}
Feed-forward methods replace per-scene optimization with a single network trained across multi-scenes. Generalizable novel view synthesis (NVS) and 3DGS models such as MVSplat \cite{chen2024mvsplat} and AnySplat \cite{jiang2025anysplat} infer geometry or novel views in one forward pass, and recent work shows that 3D Gaussians and camera poses can be predicted directly from unconstrained views. Less3Depend \cite{wang2025less} further synthesizes novel views from sparse, unposed inputs without explicit 3D or pose supervision, while 3D foundation models such as VGGT \cite{wang2025vggt} predict cameras, depth and dense point maps for dozens of views and serve as versatile backbones for downstream 3D tasks. In parallel, diffusion-based generative pipelines provide additional 3D priors. Multi-view diffusion models like SyncDreamer and Zero123++ enforce cross-view consistency from very few real images \cite{liu2023syncdreamer,shi2023zero123++,kong2025generative}, and PartGen combines two multi-view diffusion models for part-level 3D asset generation and reconstruction \cite{chen2025partgen}. These methods excel at fast and scalable novel view synthesis while they typically do not predict an explicit 3D representation or cannot iteratively optimize scene-specific models. When meeting severely sparse or occluded input views, its geometric fidelity and editability are still relatively poor, and they do not directly target high-quality per-scene 3DGS in our setting. We therefore introduce them primarily as conceptual context and potential sources of priors, and restrict our quantitative comparisons to per-scene baselines.

\subsection{Reliability-aware and Structural Supervision}
Our design is also related to reliability-aware supervision and structure-aware density control. NeRF-based works use semantic features, appearance embeddings, uncertainty estimates and view selection to downweight ambiguous observations \cite{jain2021putting,martin2021nerf,xiao2024nerf,ren2024nerf,wu2026adh,xu2026optimized,liu2025detection,kang2026machine,zhang2026highly,liu2024explicitly,liu2023robust,huang2023dense,liu2025seeing}, and diffusion or multi-view generative models such as DreamGaussian, SyncDreamer and Zero123++ provide strong priors for synthesized views but generally treat them as equally reliable, so local artifacts can still corrupt 3D optimization \cite{tang2023dreamgaussian,liu2023syncdreamer,shi2023zero123++,kong2025generative}. In explicit 3DGS, standard splatting uses view-space residuals to drive split and grow, and SLAM3R \cite{liu2025slam3r} improves robustness by aligning local point maps in real time but targets online tracking rather than offline sparse-view 3DGS. These methods underline the need for robust supervision and structure-aware densification, but they do not explicitly learn per-pixel or per-view reliability for synthesized pseudo-views or tie densification to their consistency with the evolving geometry, whereas we combine 2D-assisted perceptual enhancement, a learnable confidence module and normal-guided densification so that both supervision and Gaussian growth follow the current 3D structure.

\section{Method}
In this section, we present Hierarchical Gaussian Fields (HierGF), an explicit neural rendering framework that tackles sparse-view reconstruction through a two-stage backbone for hierarchical geometry–perception message passing. Section~\ref{motivation} outlines the motivation and the specific challenges posed by sparse inputs. Section~\ref{define} formalizes the problem and introduces the 3D Gaussian representation. Section~\ref{backbone} describes the proposed hierarchical geometry–perception backbone, in which a coarse-scale Gaussian field and a fine-scale Gaussian field are coupled via geometry–perception message passing. Section~\ref{perception} details the 2D-assisted perceptual supervision module that enhances and filters pseudo-views before they are used as training signals. Finally, Section~\ref{GCDC} introduces the geometry-consistent densification module, which performs normal-weighted, structure-aware densification to improve reconstruction in under-sampled regions.

\subsection{The Motivation}
\label{motivation}
Sparse views contain only limited 3D information and lead to two related problems. Too few images are available for matching, many regions are never seen from diverse viewpoints, reprojection losses provide little gradient, and reconstructions are easily dominated by single views with view dependent artifacts. View coverage is also poor, so large parts of the object are under-sampled and existing methods fall back on heuristic density control that often blurs missing content or produces holes, floaters and broken thin structures. 

HierGF addresses these issues by coupling geometry and perception in a hierarchical way. A coarse-scale Gaussian field is first estimated from sparse inputs to capture the main geometry, then used to render pseudo-views that are refined by a 2D-assisted perceptual supervision module with a diffusion prior and a confidence network that selects reliable pixels and views. In parallel, a geometry-consistent density completion strategy inserts new Gaussians along estimated surface directions using normal-weighted interpolation, so that a small and uneven set of views is transformed into richer supervision and under-sampled regions are filled with geometrically coherent detail.


\subsection{Problem Formulation and Representation}
\label{define}

We start with sparse reference images $\mathbf{X}^{\mathrm{ref}}=\{\mathbf{x}_i\}_{i=1}^N$ of a single object and their camera parameters $\mathbf{\Pi}^{\mathrm{ref}}=\{\mathbf{\pi}_i\}_{i=1}^N$. The goal is to learn a representation that supports photorealistic rendering from arbitrary viewpoints with coherent geometry and cross-view appearance under few input views. We adopt 3D Gaussian Splatting~\cite{kerbl20233d}, where each primitive 
is parameterized by center $\mu_j\in\mathbb{R}^3$, rotation $q_j$, anisotropic scale $s_j$, opacity $\sigma_j$, and SH color coefficients $c_j$. Rendering integrates projected anisotropic Gaussians along camera rays to synthesize images (and optionally depths). This explicit parameterization enables real-time rendering, and fine-grained density control, which are beneficial in sparse-view settings.

In the sparse-view setting, many regions are unobserved or only weakly constrained, so Gaussian density lacks a reliable update direction. HierGF addresses this with a hierarchical geometry–perception backbone: a coarse-scale field learned from sparse inputs renders additional views that are enhanced with 2D-assisted perceptual supervision and confidence weighting, then combined with geometry-consistent densification to refine a fine-scale field in under-sampled regions. This feedback gradually improves geometry and appearance and strengthens multi-view consistency, as illustrated in Fig.~\ref{fig:architecture}.

\subsection{Hierarchical Geometry–perception Message Passing}
\label{backbone}

Under sparse views, many regions are weakly constrained, so photometric losses give little guidance and density growth lacks direction. To avoid hallucinated details from early strong priors and unresolved holes from plain densification, we adopt a two-stage scheme: a coarse Gaussian field is first learned from sparse inputs, then refined into a fine-scale field using 2D-assisted, confidence-weighted supervision and geometry-consistent densification in under-sampled regions. The specific details are as follows:

\noindent \textbf{Coarse-grained rendering in the first stage.} 
The first stage aims to obtain a stable geometric backbone from very sparse observations, rather than to directly chase photo-realistic detail. Under few-view supervision, trying to fit fine texture too early often leads to overfitting to individual images and unstable density growth. We therefore focus on building a coarse but reliable Gaussian field that can be rendered consistently in the reference views and in their local neighborhoods. In this stage, the optimization emphasizes alignment of visible regions, reasonable global scale, and basic cross-view coherence, so that the resulting geometry is interpretable and can safely support subsequent perceptual enhancement and densification. Specifically, we obtain a coarse Gaussian field: 
\begin{equation}
    \mathbf{\mathcal{P}}_c=\mathcal{G}(\mathbf{x}_i,\mathbf{\pi}_i)_{i=1}^{N}.
\end{equation} Each 3D Gaussian point consists of a center position $\mu$, a rotation quaternion $q$, a scale vector $s$, an opacity $\sigma$, and spherical harmonic (SH) coefficients $sh$. The coarse Gaussian field is parameterized as a set of Gaussian points: 
\begin{equation}
    \mathcal{P}_c=\{\mathbf{p_i}:\mu_i, q_i, s_i, \sigma_i, sh_i \}_{i=1}^{P}.
\end{equation}

\noindent \textbf{Fine-grained feedback in the second stage.} 
The second stage refines the reconstruction by coupling the coarse geometry with 2D-assisted perceptual supervision. Starting from the coarse Gaussian field, we periodically render pseudo-views, enhance them with a lightweight diffusion prior to sharpen texture and harmonize shading, and feed both coarse and enhanced images into a confidence network that predicts per-pixel and per-view weights. These confidences reweight color and perceptual losses for the fine-scale field, while a geometry-consistent densification module inserts new Gaussians along estimated surface normals via normal-weighted interpolation to fill under-sampled regions in a structure-aware manner.

To avoid locking in suboptimal appearance, the fine-scale field transfers only spatial means $\mu_i$ and RGB initialization from the coarse field, while opacity, anisotropic scale, rotation, and higher-order appearance attributes are reinitialized. As training alternates between pseudo-view generation, LoRA-adapted enhancement, confidence estimation, confidence-weighted optimization, and densification, pseudo-observations track the evolving geometry and provide increasingly reliable multi-view supervision, which strengthens consistency and gradually completes previously under-constrained areas.

The confidence module and geometry-consistent densification play different roles in refining the fine field: confidence weighting reweights the losses associated with enhanced pseudo-views, whereas geometry-consistent densification modifies the Gaussian representation in sparse regions according to local geometric consistency. The confidence weights are not used to control the densification rule.

Table~\ref{tab:r1_c3_parameter_data} further specifies, for each pipeline stage, which data are used, which parameters are updated, and which components remain fixed, making explicit that the diffusion backbone is not jointly optimized with the final Gaussian representation.


\begin{table*}[t]
\centering
\caption{Parameter status and data usage in the diffusion-assisted reconstruction pipeline.}
\label{tab:r1_c3_parameter_data}
\small
\setlength{\tabcolsep}{3.2pt}
\setlength{\extrarowheight}{1pt}
\renewcommand{\arraystretch}{1.28}
\begin{tabular}{@{}
>{\raggedright\arraybackslash}p{0.16\textwidth}
>{\raggedright\arraybackslash}p{0.18\textwidth}
>{\raggedright\arraybackslash}p{0.15\textwidth}
>{\raggedright\arraybackslash}p{0.22\textwidth}
>{\raggedright\arraybackslash}p{0.20\textwidth}
@{}}
\hline
\textbf{Component / stage} &
\textbf{Data used} &
\textbf{Updated parameters} &
\textbf{Fixed parameters} &
\textbf{Output} \\
\hline
Coarse Gaussian optimization &
Sparse reference views &
Coarse Gaussians &
Diffusion module not used &
Coarse Gaussian field \\
\hline
Pair construction &
Reference views; perturbed coarse renderings &
None &
Optimized coarse Gaussians &
Degraded/reference image pairs \\
\hline
LoRA adaptation &
Degraded renderings and reference images &
LoRA only &
Pre-trained ControlNet/\allowbreak SD v1.5 backbone &
Scene-adapted enhancer \\
\hline
Pseudo-view rendering &
Additional camera poses &
None &
Coarse Gaussian field &
Coarse pseudo-views \\
\hline
Pseudo-view enhancement &
Coarse pseudo-views &
None &
Diffusion backbone and adapted LoRA &
Enhanced pseudo-\allowbreak supervision \\
\hline
Fine-scale optimization &
Reference views; enhanced pseudo-\allowbreak supervision &
Fine Gaussians; confidence module &
Diffusion backbone and adapted LoRA &
Final Gaussian representation \\
\hline
Novel-view rendering &
Novel camera poses &
None &
All optimized parameters &
Rendered views; no diffusion use \\
\hline
\end{tabular}
\vspace{-4mm}
\end{table*}

\subsection{2D-assisted Perceptual Enhancement}
\label{perception}
In sparse-view reconstruction, raw reprojection from a few images provides weak and uneven constraints, which easily leads to blurry textures and inconsistent shading. Our 2D-assisted perceptual enhancement addresses this by rendering extra views from the coarse-scale Gaussian field, enhancing them with a diffusion-based 2D prior, and then filtering them with a confidence network that predicts pixel-level and view-level weights. These weights are used only to reweight the color and perceptual losses, so the fine-scale Gaussian field is trained under enhanced yet selective supervision that strengthens geometrically consistent, cross-view stable content while suppressing unreliable regions.

\noindent \textbf{Self-generated perceptual supervision.} 
Due to incomplete perspective coverage, 3D Gaussians learned from sparse views often produce blurry or inconsistent regions. A pre-trained 2D prior, however, can restore missing details and yield more realistic images, even for coarse multi-view renderings. Motivated by this, we follow the per-scene enhancement strategy and inject 3D noise $\mathbf{\epsilon}_s$ into Gaussian attributes to render degraded/reference pairs from the current scene. We then adapt only the LoRA~\cite{hu2022lora} parameters of a ControlNet~\cite{zhang2023controlnet} enhancement module based on Stable Diffusion v1.5~\cite{rombach2022latent}; the diffusion backbone remains frozen throughout this adaptation and during reconstruction.

Next, from the coarse Gaussian field trained in the first stage, we obtain $N^{\prime}$ rendered images $\mathbf{X}^{\mathrm{c}}=\{\mathbf{x}_j\}_{j=1}^{N^{\prime}}$ from new viewpoints.
The coarse image $\mathbf{X}^{c}$ is fed into the fine-tuned 2D prior model to obtain the restored image $\Tilde{\mathbf{X}}^{c}$. The enhanced reference image is denoted as: 
\begin{equation}
    \mathbf{X}^{\mathrm{enh}}=\mathbf{X}^{\mathrm{ref}}\cup \Tilde{\mathbf{X}}^{\mathrm{c}},
\end{equation} and the corresponding camera parameters are denoted as: 
\begin{equation}
    \mathbf{\Pi}^{\mathrm{enh}}=\mathbf{\Pi}^{\mathrm{ref}}\cup \mathbf{\Pi}^{c}.
\end{equation}
We use this as a self-supervisory signal to optimize the fine Gaussian field in the second stage to obtain photorealistic renderings from any viewpoint, denoted as: 
\begin{equation}
    \mathbf{x} = \mathcal{G}_f(\pi|{\mathbf{x}_i, \mathbf{\pi}_i}_{i=1}^{N+N^{\prime}}).
\end{equation}
By avoiding the inclusion of extraneous visual information, our method ensures that all refinement operations are based solely on the provided dataset, preserving the original conditions and maximizing the use of available data. This independent refinement mechanism distinguishes our approach and is crucial for achieving highly accurate and consistent 3D models.

The optimization of both the first-stage Gaussian function $\mathcal{G}_c$ and the second-stage Gaussian function $\mathcal{G}_f$ incorporates color and monocular depth losses. The color loss function combines the L1 loss function and the D-SSIM loss function:
\begin{equation}
\mathbf{\mathcal{L}}_1=||\mathbf{x}-\mathbf{x}^{\mathrm{ref}}||_{1}, \mathcal{L}_{\mathrm{D-SSIM}}=1-\mathrm{SSIM}(\mathbf{x},\mathbf{x}^{\mathrm{ref}}),
\end{equation}
where $x$ is the rendered image and $x^{ref}$ is the corresponding reference image.
A translation- and scale-invariant depth loss is used to guide geometry:
\begin{equation}
\mathcal{L}_{\mathrm{d}}=||\mathbf{D}^*-\mathbf{D}_{\mathrm{pred}}^*||_{1},
\end{equation}
where $\mathbf{D}^*$ and $\mathbf{D}^*_{\mathrm{pred}}$ are the rendered depth and the monocular estimated depth per frame, respectively, \cite{bhat2023zoedepth}. Depth values are computed using the normalization strategy \cite{ranftl2020robust}:
\begin{equation}
\mathbf{D}^*=\frac{\mathbf{D}-\mathrm{median}(\mathbf{D})}{\frac{1}{M}\Sigma_{i=1}^M|\mathbf{D}-\mathrm{median}(\mathbf{D})|},
\end{equation}
where $M$ represents the number of valid pixels. The overall loss function combines these components as follows:
\begin{equation}
\mathcal{L}_{\mathrm{gs}}=(1-\lambda_{\mathrm{SSIM}})\mathcal{L}_1 +\lambda_{\mathrm{SSIM}}\mathcal{L}_{\mathrm{D-SSIM}}+\lambda_{\mathrm{d}}\mathcal{L}_{\mathrm{d}},
\end{equation}
where $\lambda_{\mathrm{SSIM}}$ and $\lambda_{\mathrm{d}}$ control the magnitude of each term.

\noindent \textbf{Confidence-guided supervision.} 
Some pseudo-observations generated by the 2D prior are unreliable in sparse-view settings, especially in occluded areas, along thin structures, or at grazing angles, and directly using them can turn enhancement artifacts into wrong geometry. To avoid this, we introduce a lightweight confidence network that takes each coarse rendering and its enhanced counterpart and predicts both a pixel-wise confidence map and a view-wise weight, indicating which regions and views are consistent with the current geometry. These confidences are used only to reweight the reprojection and perceptual losses, so that gradients focus on reliable pixels and views while unstable regions have reduced influence.

Specifically, given the $j$-th pseudo-view, let $\mathbf{x}_j^c$ denote the coarse rendering and $\tilde{\mathbf{x}}_{j}^{c}$ its perceptual enhancement. We form the input tensor as:
\begin{equation}
    \mathbf{I}_j=[\mathbf{x}_j^{c},\tilde{\mathbf{x}}_j^{c},|\mathbf{x}_j^{c}-\tilde{\mathbf{x}}_j^{c}|],
\end{equation}
We pass $\mathbf{I}_j$ through a small convolutional encoder $E_{\theta}$ to obtain a shared feature map $\mathbf{F}_j=E_{\theta}(\mathbf{I}_j)$. The pixel head $H_{\mathrm{pix}}$ upsamples and decodes $\mathbf{F}_j$ with two stages of $3\times3$ convolutions and outputs a single-channel confidence map:
\begin{equation}
    \mathbf{M}_j = \sigma \big(H_{\mathrm{pix}}(\mathbf{F}_j)\big),
\end{equation}
where $\sigma$ denotes the sigmoid function. To avoid instability from global scaling, we apply per-image mean normalization so that the pixel weights only redistribute mass within the image:
\begin{equation}
    \hat{\mathbf{M}}_j(p)=\frac{\mathbf{M}_j(p)}{\frac{1}{HW}\sum\nolimits_{q}\mathbf{M}_j(q)},\qquad
\frac{1}{HW}\sum\nolimits_{p}\hat{\mathbf{M}}_j(p)=1,
\end{equation}
where $H$ and $W$ represent the image height and width, respectively. The pixel indices $p$ and $q$ range over the $H\times W$ spatial grid.
The view head $H_{\mathrm{view}}$ first obtains $\bar{\mathbf{F}}_j$ via global average pooling on $\mathbf{F}_j$, and then concatenates it with the global information of the three input images to get: 
\begin{equation}
    \mathbf{g}_j = \big[\bar{\mathbf{F}}_j,\mu(\mathbf{x}_j^{c}),\mu(\tilde{\mathbf{x}}_j^{c}),\mu(|\mathbf{x}_j^{c}-\tilde{\mathbf{x}}_j^{c}|)\big],
\end{equation}
where $\mu(\cdot)$ denotes the per-channel spatial mean. A two-layer MLP followed by sigmoid function outputs the view confidence:
\begin{equation}
    \mathbf{w}_j=\sigma(H_{\mathrm{view}}(\mathbf{g}_j))\in(0,1).
\end{equation}

Let $\mathbf{x}(\pi^{\prime}_j)$ be the current rendering at pose $\pi_j^\prime$, and define the supervision target as:
\begin{equation}
    \mathbf{y}_j = \begin{cases} \mathbf{x}_j^{\mathrm{ref}}, & \pi_j^{\prime} \in \Pi^{\mathrm{ref}} \\[2pt] \tilde{\mathbf{x}}_j^{c}, & \pi_j^{\prime} \in \Pi^{\mathrm{c}} \end{cases},
\end{equation}
we reweight only the $\mathcal{L}_{\mathrm{1}}$ and $\mathcal{L}_{\mathrm{D-SSIM}}$ terms with the learned confidences, keeping all other terms unchanged:
\begin{align}
    &\mathcal{L}_{1}^{w}
=\sum_{j\in N^\prime } \mathbf{w}_j \sum_{p}\hat{\mathbf{M}}_j(p)\big|\mathbf{x}(\pi^{\prime}_j)^{p}-\mathbf{y}_j^{p}\big|
, \\ &\mathcal{L}_{\mathrm{D-SSIM}}^{w}
=\sum_{j\in N^\prime } \mathbf{w}_j \big( 1-\mathrm{SSIM}(\mathbf{x}(\pi^{\prime}_j),\mathbf{y}_j) \big),
\end{align}
and the overall objective replaces the original color terms by their weighted counterparts as follows:
\begin{equation}
\mathcal{L}_{\mathrm{gs}}^w=(1-\lambda_{\mathrm{SSIM}})\mathcal{L}_1^w +\lambda_{\mathrm{SSIM}}\mathcal{L}_{\mathrm{D-SSIM}}^w+\lambda_{\mathrm{d}}\mathcal{L}_{\mathrm{d}}.
\end{equation}

\begin{table*}
\caption{Comparisons on the MipNeRF360 dataset with varying input views. LPIPS$^{*}$ = LPIPS x \(10^2\) and SSIM$^{*}$ = SSIM x \(10^2\) throughout the paper.}
\label{tab:table1}
\centering
\resizebox{0.9\textwidth}{!}{
\begin{tabular}{@{}c@{}c@{}cccccccccc@{}}
\toprule
& & \multirow{2.5}{*}{\textbf{Method}} & \multicolumn{3}{c}{\textbf{4-view}} & \multicolumn{3}{c}{\textbf{6-view}} & \multicolumn{3}{c}{\textbf{9-view}} \\ 
\cmidrule(r){4-6} \cmidrule(lr){7-9} \cmidrule(l){10-12}
& & & LPIPS$^{*}$ ↓ & PSNR ↑ & SSIM$^{*}$ ↑ & LPIPS$^{*}$ ↓ & PSNR ↑ & SSIM$^{*}$ ↑ & LPIPS$^{*}$ ↓ & PSNR ↑ & SSIM$^{*}$ ↑ \\
\midrule
\multirow{10}{*} & & DVGO \cite{sun2022direct} & 24.43 & 14.39 & 79.12 & 26.67 & 14.30 & 76.76 & 25.66 & 14.74 & 78.42 \\
& & 3DGS \cite{kerbl20233d} & 10.80 & 20.31 & 89.91 & 8.38 & 22.12 & 91.34 & 6.42 & 24.29 & 93.31 \\
& & DietNeRF \cite{jain2021putting} & 11.17 & 18.90 & 89.71 & 6.96 & 22.03 & 92.86 & 5.85 & 23.55 & 94.24 \\
& & RegNeRF \cite{Niemeyer2021Regnerf} & 20.44 & 13.59 & 84.76 & 20.72 & 13.41 & 84.18 & 19.70 & 13.68 & 85.17 \\
& & FreeNeRF \cite{10205351} & 16.83 & 13.71 & 85.34 & 6.84 & 22.26 & 93.32 & 5.51 & 27.66 & 94.85 \\
& & SparseNeRF \cite{wang2023sparsenerf} & 17.76 & 12.83 & 84.54 & 19.74 & 13.42 & 83.16 & 21.56 & 14.36 & 82.35 \\
& & ZeroRF \cite{shi2024zerorf} & 19.88 & 14.17 & 81.88 & 8.31 & 24.14 & 92.11 & 5.34 & 27.78 & 94.60 \\
& & FSGS \cite{zhu2023FSGS} & 9.51 & 21.07 & 90.97 & 7.69 & 22.68 & 92.64 & 6.06 & 25.31 & 93.97 \\
& & GS-GS \cite{kong2025generative} & 7.65   &  22.39  & 91.87   & 5.71   & 25.87   &  94.72  &  4.58  &  27.30  & 94.26   \\
& & GaussianObject \cite{yang2024gaussianobject} & 4.98 & 24.81 & 93.50 & 3.63 & 27.00 & 95.12 & 2.75 & 28.62 & 96.38 \\
& & WaveletGS \cite{nguyen2025waveletgaussian} & 4.70  & 25.31   &  93.91  & 3.52   &  28.16  & 95.78   & 2.72   & 28.97   & 96.80   \\
\cmidrule{2-12}
& & HierGF (ours)  & \textbf{3.85} & \textbf{28.48} & \textbf{95.87} & \textbf{3.14} & \textbf{30.02} & \textbf{96.85} & \textbf{2.58} & \textbf{30.96} & \textbf{97.32} \\

\bottomrule
\end{tabular}}
\end{table*}

\begin{table*}
\caption{Comparisons on the OmniObject3D datasets with varying input views. }
\label{tab:table2}
\centering
\resizebox{0.9\textwidth}{!}{
\begin{tabular}{@{}c@{}c@{}cccccccccc@{}}
\toprule
& & \multirow{2.5}{*}{\textbf{Method}} & \multicolumn{3}{c}{\textbf{4-view}} & \multicolumn{3}{c}{\textbf{6-view}} & \multicolumn{3}{c}{\textbf{9-view}} \\ 
\cmidrule(r){4-6} \cmidrule(lr){7-9} \cmidrule(l){10-12}
& & & LPIPS$^{*}$ ↓ & PSNR ↑ & SSIM$^{*}$ ↑ & LPIPS$^{*}$ ↓ & PSNR ↑ & SSIM$^{*}$ ↑ & LPIPS* ↓ & PSNR ↑ & SSIM$^{*}$ ↑ \\
\midrule

\multirow{10}{*} & & DVGO \cite{sun2022direct} & 14.48 & 17.14 & 89.52 & 12.89 & 18.32 & 91.42 & 11.49 & 19.26 & 93.02 \\
& & 3DGS \cite{kerbl20233d} & 8.60 & 17.29 & 92.99 & 7.74 & 18.29 & 93.78 & 6.50 & 20.26 & 94.83 \\
& & DietNeRF \cite{jain2021putting} & 11.64 & 18.56 & 92.05 & 10.39 & 19.07 & 92.67 & 10.32 & 19.26 & 92.58 \\
& & RegNeRF \cite{Niemeyer2021Regnerf} & 16.75 & 15.20 & 90.91 & 14.38 & 15.80 & 92.07 & 10.17 & 17.93 & 94.20 \\
& & FreeNeRF \cite{10205351} & 8.28 & 17.78 & 94.02 & 7.32 & 19.02 & 94.64 & 7.25 & 20.35 & 94.67 \\
& & SparseNeRF \cite{wang2023sparsenerf} & 17.47 & 15.22 & 89.21 & 21.71 & 15.86 & 89.35 & 23.76 & 17.16 & 89.47 \\
& & ZeroRF \cite{shi2024zerorf} & 4.44 & 27.78 & 96.15 & 3.11 & 31.94 & 97.31 & 3.10 & 32.93 & 97.47 \\
& & FSGS \cite{zhu2023FSGS} & 6.25 & 24.71 & 95.45 & 6.05 & 26.36 & 95.82 & 4.17 & 29.16 & 96.95 \\
& & GaussianObject \cite{yang2024gaussianobject} & 2.07 & 30.89 & 97.56 & 1.55 & 33.31 & 98.21 & 1.20 & 35.49 & 98.70 \\
& & WaveletGS \cite{nguyen2025waveletgaussian} & 2.80 & 31.22  & 98.30 & 1.58 & 33.64 & 98.36 & 1.26 & 35.60 &  98.78 \\
\cmidrule{2-12}
& & HierGF (ours) & \textbf{1.76} & \textbf{34.08} & \textbf{98.36} & \textbf{1.48} & \textbf{34.25} & \textbf{99.14} & \textbf{1.12} & \textbf{35.88} & \textbf{99.30} \\
\bottomrule
\end{tabular}}
\vspace{-3mm}
\end{table*}

Furthermore, considering that relying solely on the weighted reconstruction loss to update the confidence network can easily lead to weight collapse to all zeros or all ones, affecting training stability, we introduce a simple regularization term to constrain the distribution of confidence scores as follows:
\begin{align}
    \mathcal{L}_{\mathrm{conf}}
    &= \lambda_{\mathrm{view}} \sum_{j\in N\prime} \big(\mathbf{w}_j - 1\big)^{2}
    + \lambda_{\mathrm{tv}} \sum_{j\in N\prime} \operatorname{TV}(\mathbf{M}_j)  \nonumber \\
    &+ \lambda_{\mathrm{pix}}\sum_{j\in N\prime} \Big(\overline{\mathbf{M}}_j - 1\Big)^{2},
\end{align}
where $\overline{\mathbf{M}}_j = \frac{1}{HW}\sum_{p} \mathbf{M}_j(p)$, $\operatorname{TV}(\cdot)$ denotes a standard total variation regularizer, and $\lambda_{\mathrm{pix}}$, $\lambda_{\mathrm{view}}$, and $\lambda_{\mathrm{tv}}$ are small positive weights. The full objective is:
\begin{equation}
    \mathcal{L}_{\mathrm{total}}
    = \mathcal{L}_{\mathrm{gs}}^{w}
      + \mathcal{L}_{\mathrm{conf}}.
\end{equation}

\subsection{Geometry-Consistent Densification}
\label{GCDC}


Beyond providing extra view information through perceptual enhancement, we further refine Gaussian optimization in the refined field to improve geometric continuity. In 3DGS-based reconstruction \cite{kerbl20233d}, traditional Adaptive Density Control (ADC) adjusts density from local sparsity and inter-view consistency but often fails in sparse regions with weak gradients, where naive interpolation from few neighbors produces gaps or distortions, especially in complex or very low-density areas. To address this, we introduce a normal-weighted interpolation strategy that uses neighboring normals to place new Gaussians along the underlying surface, improving detail completion while preserving local structure.

Specifically, we voxelize the 3D space into fixed-size cells and count the number of Gaussians in each voxel. Voxels with fewer than a threshold $\tau$ points are marked as sparse, so optimization can concentrate on these regions and avoid unnecessary computation. Let $D_{\mathrm{rob}}$ denote the robust scene extent, computed as the 95th percentile of pairwise distances among Gaussian centers after removing extreme outliers. We set the voxel size to
\[
s_v = 0.02D_{\mathrm{rob}},
\]
and use the default neighborhood radius
\[
r=2.0s_v=0.04D_{\mathrm{rob}}.
\]
For stable PCA-based normal estimation, at least $K_{\min}=8$ neighboring Gaussians are required and at most the nearest $K_{\max}=32$ neighbors are retained. If fewer than eight neighbors are found for a seed Gaussian, normal-guided insertion is skipped for this seed and the original ADC update is retained. For each point $\mathbf{p}_i$ in a sparse voxel, we define a radius-$r$ neighborhood $\mathcal{N}(\mathbf{p}_i, r)$ and estimate its normal $\mathbf{n}_i$ by applying PCA to the neighboring points. For each neighboring point $\mathbf{p}_j\in \mathcal{N}(\mathbf{p}_i, r)$, a weighting function is defined based on the similarity in normal direction and spatial distance. The weight based on normal similarity is given by:
\begin{equation}
    \mathbf{w}_{\text{normal}} = \exp(-\alpha \|\mathbf{n}_i - \mathbf{n}_j\|^2),
\end{equation}
where $\mathbf{n}_j $ is the normal of neighboring point \( \mathbf{p}_j \), and \(\alpha\) is a hyperparameter controlling the influence of normal similarity. The spatial distance weight is defined as:

\begin{equation}
    \mathbf{w}_{\text{distance}} = \exp(-\beta \|\mathbf{p}_i - \mathbf{p}_j\|^2),
\end{equation}
where $\beta$ is a hyperparameter controlling the impact of distance. The final weight for interpolation between $ \mathbf{p}_i$ and $\mathbf{p}_j$ is then defined as the product of these two weights:
\begin{equation}
\mathbf{w}_{ij} = \mathbf{w}_{\text{normal}} \cdot \mathbf{w}_{\text{distance}}.
\end{equation}

Using these weights, we perform a weighted average of the positions and attributes of neighboring points to compute the interpolated point's position and attributes. This interpolation operation is applied to every point in the sparse region, resulting in an interpolated region with twice the original density. This normal-based weighted interpolation strategy effectively preserves the geometric consistency within sparse voxels, generating new points that conform to the original geometric structure and mitigate distortions that may arise from arbitrary interpolation.

\section{Experiments}

\begin{figure*}
  \centering
  \vspace{-5mm}
  \includegraphics[width=0.9\textwidth]{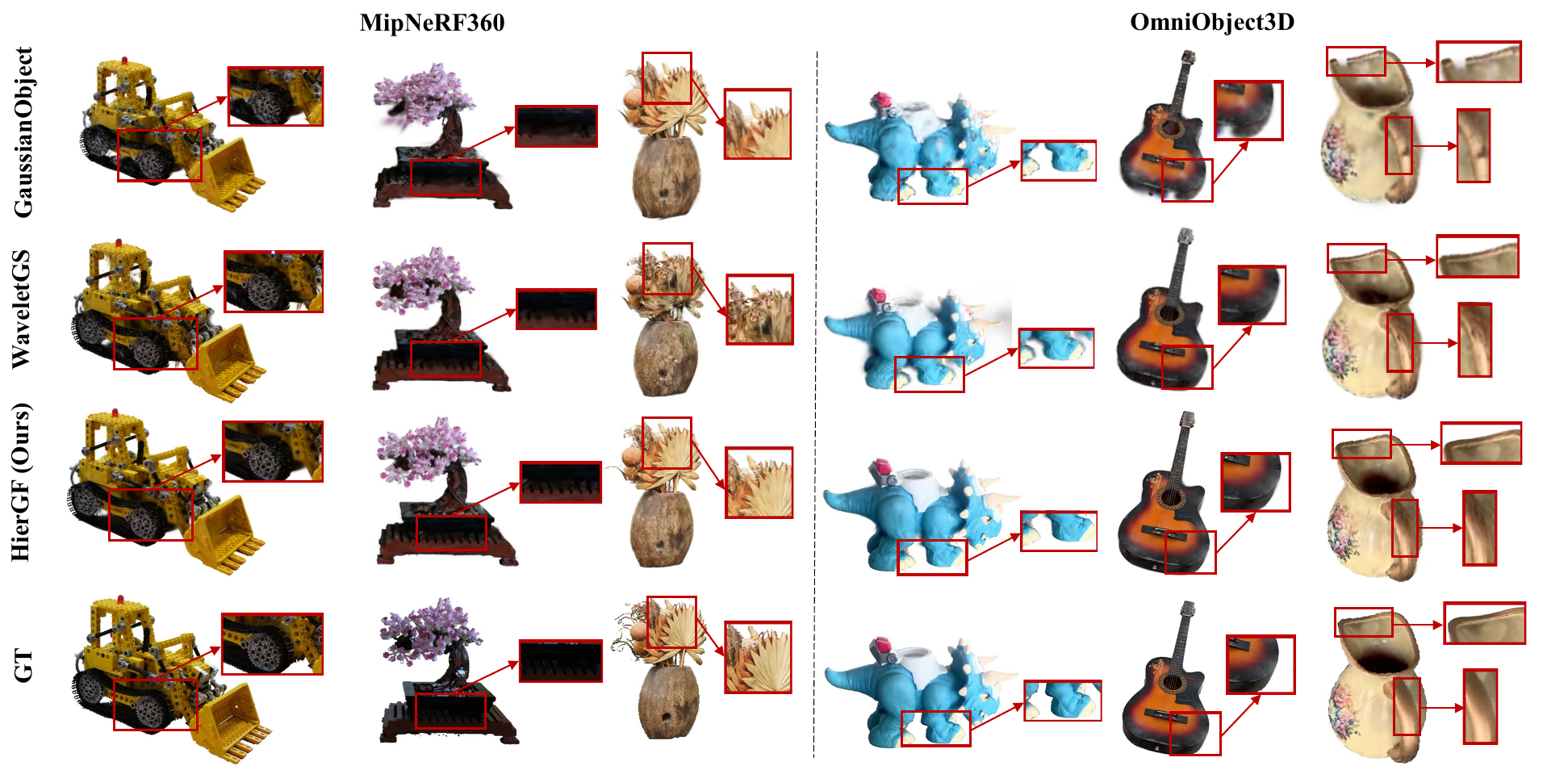}
  \caption{Qualitative examples on the MipNeRF360 and OmniObject3D datasets with 4 input views.}
  
  \label{fig:vis1}
  \vspace{-4mm}
\end{figure*}

\begin{figure*}
  \centering
  \includegraphics[width=0.9\textwidth]{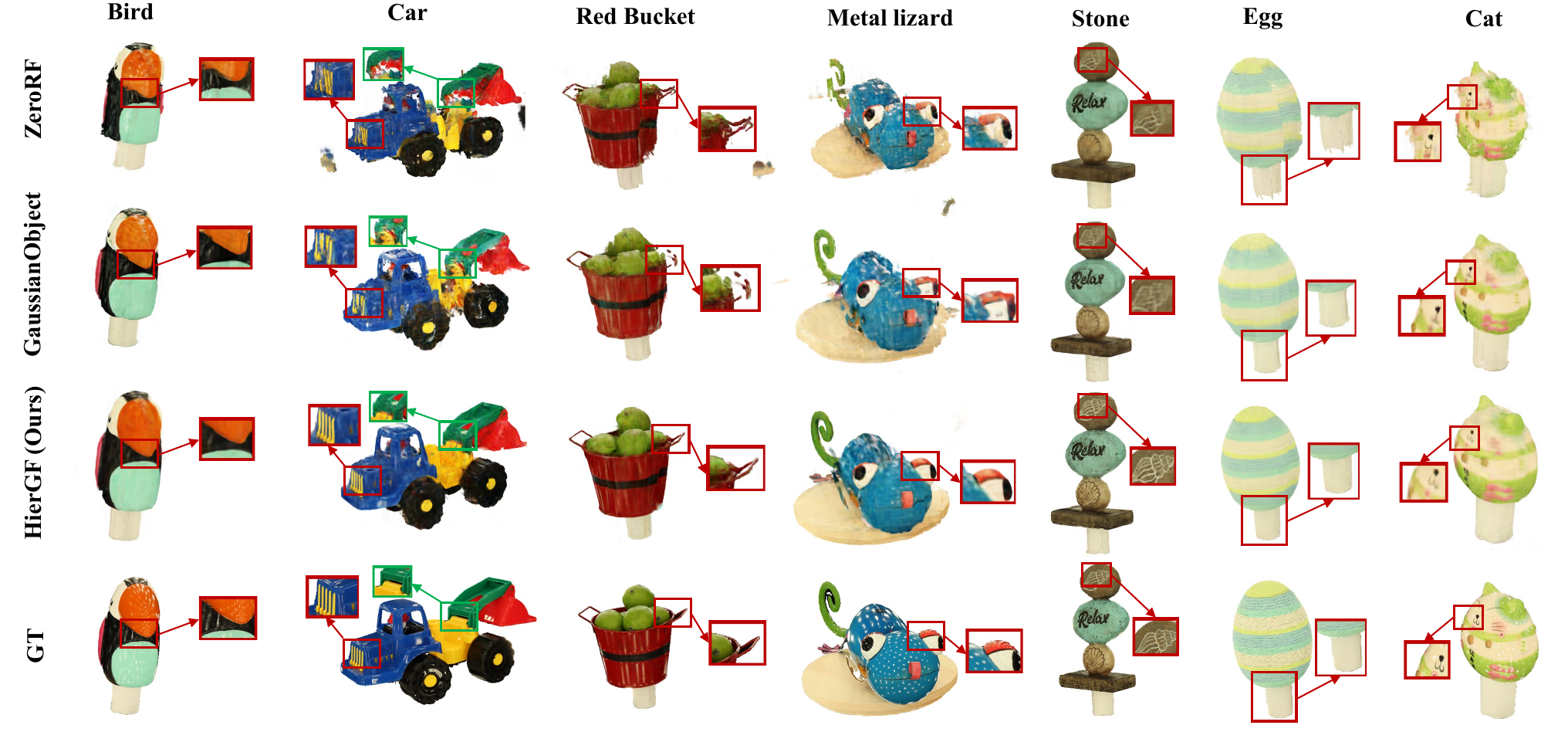}
  \caption{Qualitative examples on the OpenIllumination dataset with 4 input views.}
  
  \label{fig:open}
\vspace{-3mm}
\end{figure*}

\subsection{Implementation Detail}
Our reconstruction framework extends 3D Gaussian Splatting (3DGS)~\cite{kerbl20233d} and casts sparse-view reconstruction as two-stage hierarchical geometry-perception feedback. The coarse field is optimized for \(4{,}000\) iterations, with standard adaptive density control every \(100\) iterations from iteration \(500\) to \(3{,}500\) and opacity reset at iterations \(1{,}500\) and \(3{,}000\). This stage uses ZoeDepth~\cite{bhat2023zoedepth} as translation- and scale-invariant monocular depth supervision. After coarse reconstruction, degraded/reference image pairs are constructed from the input views and only the LoRA parameters of the ControlNet / Stable Diffusion v1.5 enhancement module are adapted for \(500\) optimization steps, while the diffusion backbone remains frozen. The fine Gaussian field then inherits only spatial means and RGB initialization from the coarse field and is optimized for \(6{,}000\) iterations. Enhanced pseudo-views are refreshed every \(K=500\) fine-stage iterations, and geometry-consistent densification is applied during the fine stage. The total Gaussian optimization budget is \(10{,}000\) iterations per scene, and the measured per-scene reconstruction time is approximately 10.5 minutes (632~s) at \(779\times520\) on a single NVIDIA A100 GPU.

The resulting implementation and evaluation settings are summarized in Table~\ref{tab:r2_c3_hiergf_settings}, including camera usage, SfM initialization, optimization budgets, LoRA adaptation, pseudo-view refresh, input resolution, and hardware.

\begin{table*}[t]
\centering
\caption{Implementation and evaluation settings of HierGF.}
\label{tab:r2_c3_hiergf_settings}
\scriptsize
\setlength{\tabcolsep}{4pt}
\renewcommand{\arraystretch}{1.15}
\resizebox{\textwidth}{!}{%
\begin{tabular}{p{4.2cm}p{11.2cm}}
\hline
\textbf{Item} & \textbf{Setting} \\
\hline
Representation &
Per-scene optimization with a coarse Gaussian field and a separately optimized fine Gaussian field \\
\hline
Camera parameters &
Official benchmark intrinsics and extrinsics; fixed during optimization; no camera-pose refinement \\
\hline
SfM initialization input &
Selected sparse input views only; no held-out evaluation views \\
\hline
SfM initialization procedure &
Sparse feature extraction and exhaustive matching among input views, followed by triangulation under fixed benchmark cameras \\
\hline
SfM filtering settings &
Maximum reprojection error \(\leq 4\) pixels; minimum triangulation angle \(1.5^{\circ}\); minimum track length \(2\) views \\
\hline
Initial Gaussian attributes &
Gaussian centers initialized from triangulated SfM points; RGB initialized from observed colors in input views \\
\hline
Monocular depth prior &
ZoeDepth, used as a translation- and scale-invariant depth supervision term \\
\hline
Coarse-stage optimization &
\(4{,}000\) iterations \\
\hline
Coarse-stage density control &
Adaptive density control every \(100\) iterations from iteration \(500\) to \(3{,}500\); opacity reset at iterations \(1{,}500\) and \(3{,}000\) \\
\hline
2D enhancement prior &
ControlNet based on Stable Diffusion v1.5; diffusion backbone frozen; only LoRA parameters adapted using current-scene degraded/reference pairs \\
\hline
LoRA adaptation budget &
\(500\) optimization steps \\
\hline
Fine-field initialization &
Transfers only spatial means and RGB initialization; reinitializes opacity, anisotropic scale, rotation, and higher-order appearance attributes \\
\hline
Fine-stage optimization &
\(6{,}000\) iterations \\
\hline
Pseudo-view refresh schedule &
Every \(K=500\) fine-stage iterations \\
\hline
Total Gaussian optimization budget &
\(10{,}000\) iterations per scene: \(4{,}000\) coarse-stage iterations + \(6{,}000\) fine-stage iterations \\
\hline
Diffusion use during final rendering &
Not used; the diffusion module is used only for pseudo-view enhancement during reconstruction \\
\hline
Input resolution &
\(779\times520\) \\
\hline
Hardware &
Single NVIDIA A100 GPU \\
\hline
\end{tabular}%
}
\vspace{-4mm}
\end{table*}

\subsection{Dataset}
We evaluate our proposed method on three datasets suited for sparse-view 360$^{\circ}$ object reconstruction with varying input views, including Mip-NeRF360 \cite{Barron2021MipNeRF3U}, OmniObject3D \cite{wu2023omniobject3d}, and OpenIllumination \cite{liu2024openillumination}.
The Mip-NeRF 360 dataset addresses the synthesis challenges in unbounded scenes with a novel approach, including non-linear parameterization and online distillation.
The OmniObject3D dataset features a comprehensive collection of over 6,000 high-quality, real-scanned 3D objects across 190 daily life categories, aimed at enhancing 3D representation learning. The OpenIllumination dataset is a real-world dataset captured using LightStage. We utilize the sparse OpenIllumination dataset introduced in ZeroRF \cite{shi2024zerorf}.

\begin{table}
\centering
\caption{Quantitative comparisons on the OpenIllumination dataset. Methods with \textdagger mean the metrics are from the ZeroRF paper \cite{shi2024zerorf}. We follow ZeroRF \cite{shi2024zerorf} on dataset sampling and processing.}
\resizebox{\columnwidth}{!}{
\label{tab:table3}
\begin{tabular}{l|ccc|ccc}
\toprule
\multirow{2}{*}{\textbf{Method}} & \multicolumn{3}{c|}{\textbf{4-view}} & \multicolumn{3}{c}{\textbf{6-view}} \\ 
                & LPIPS$^{*}$ $\downarrow$ & PSNR $\uparrow$ & SSIM$^{*}$ $\uparrow$ & LPIPS$^{*}$ $\downarrow$ & PSNR $\uparrow$ & SSIM$^{*}$ $\uparrow$ \\ \hline
DVGO \cite{sun2022direct}           & 11.84           & 21.15           & 89.73          & 8.83            & 23.79           & 92.09       \\ 
3DGS \cite{kerbl2023gaussiansplatting}           & 30.08           & 11.50           & 84.54          & 29.65           & 11.98           & 82.77       \\ 
DietNeRF\textdagger \cite{jain2021putting} & 10.66       & 23.09           & 93.61         & 9.51            & 24.20           & 94.01       \\ 
RegNeRF\textdagger \cite{Niemeyer2021Regnerf} & 47.31         & 11.61           & 69.40          & 30.28           & 14.08           & 85.86      \\ 
FreeNeRF\textdagger \cite{10205351} & 35.81         & 12.21           & 79.69          & 35.15           & 11.47           & 81.28\\ 
SparseNeRF \cite{wang2023sparsenerf} & 22.28           & 13.60           & 88.08          & 26.30           & 12.80           & 84.03      \\ 
ZeroRF\textdagger \cite{shi2024zerorf} & 9.74           & 24.54           & 93.08          & 7.96            & 26.51           & 94.15      \\ 
GaussianObject \cite{yang2024gaussianobject}       & 6.71            & 24.64           & 93.54          & 5.44            & 26.54           & 94.43   \\
\midrule
HierGF (ours)        & \textbf{5.86}            & \textbf{24.66}           & \textbf{93.72}         & \textbf{5.08}            & \textbf{26.55}           & \textbf{94.53}   \\
\bottomrule
\end{tabular}
}
\vspace{-4mm}
\end{table}

\subsection{Evaluation Results \& Analysis}

In our main evaluation we compare HierGF with a broad set of sparse-view reconstruction methods on three benchmarks, as reported in Tab.~\ref{tab:table1}, Tab.~\ref{tab:table2} and Tab.~\ref{tab:table3}. The competitors include Gaussian based approaches WaveletGaussian \cite{nguyen2025waveletgaussian}, GS-GS \cite{kong2025generative} and GaussianObject \cite{yang2024gaussianobject}, vanilla 3DGS \cite{kerbl20233d} with random initialization, DVGO \cite{sun2022direct}, few-view NeRF variants RegNeRF \cite{Niemeyer2021Regnerf}, DietNeRF \cite{jain2021putting}, SparseNeRF \cite{wang2023sparsenerf} and ZeroRF \cite{shi2024zerorf}, as well as FSGS \cite{zhu2023FSGS} which builds on Gaussian splatting with SfM point initialization and extra SfM points for the highly sparse 360 degree setting.

For directly related recent sparse-view 3DGS methods, we explicitly distinguish reported results from re-run results and disclose method-specific priors or initialization assumptions. This avoids treating comparisons with different external priors as identical-configuration comparisons while keeping dataset splits, input-view protocols, evaluation views, image resolution, metric implementations, and hardware aligned whenever a re-run is used.

Table~\ref{tab:r2_c3_method_setup} lists the representation type, external priors or initialization assumptions, aligned benchmarks, and result source for each directly related sparse-view reconstruction method.

\begin{table*}[t]
\centering
\caption{Experimental assumptions of directly related sparse-view reconstruction methods.}
\label{tab:r2_c3_method_setup}
\small
\setlength{\tabcolsep}{3.2pt}
\setlength{\extrarowheight}{1pt}
\renewcommand{\arraystretch}{1.28}
\begin{tabular}{@{}
>{\raggedright\arraybackslash}p{0.15\textwidth}
>{\raggedright\arraybackslash}p{0.16\textwidth}
>{\raggedright\arraybackslash}p{0.29\textwidth}
>{\raggedright\arraybackslash}p{0.18\textwidth}
>{\raggedright\arraybackslash}p{0.17\textwidth}
@{}}
\hline
\textbf{Method} &
\textbf{Representation} &
\textbf{Prior / initialization} &
\textbf{Shared benchmarks} &
\textbf{Source / scope} \\
\hline
HierGF (ours) &
Two-field per-scene 3DGS &
Sparse SfM; ZoeDepth~\cite{bhat2023zoedepth}; LoRA-adapted ControlNet~\cite{zhang2023controlnet}/\allowbreak SD v1.5~\cite{rombach2022latent} &
MipNeRF360, OmniObject3D, OpenIllumination, LLFF, DTU &
Ours on one A100; Table~\ref{tab:r2_c3_hiergf_settings} settings \\
\hline
GaussianObject~\cite{yang2024gaussianobject} &
Per-scene Gaussian &
Visual hull; diffusion-based Gaussian repair &
Original shared benchmarks &
Reported/Re-run as marked \\
\hline
AugGS~\cite{du2024auggs} &
Two-stage Gaussian reconstruction &
Diffusion-enhanced views; structural masks &
MipNeRF360, OmniObject3D, OpenIllumination &
Related method; compatible entries marked by source \\
\hline
CoR-GS~\cite{zhang2024corgs} &
Per-scene 3DGS with co-\allowbreak regularization &
Co-pruning; pseudo-view co-\allowbreak regularization; method-specific initialization &
LLFF, DTU &
Public Reported; aligned Re-run when used \\
\hline
NexusGS~\cite{zheng2025nexusgs} &
Per-scene 3DGS &
Epipolar-depth-guided dense point-cloud initialization &
LLFF, DTU &
Public Reported; aligned Re-run when used \\
\hline
IPSM-Gaussian~\cite{wang2024diffusionpriors} &
Diffusion-guided 3DGS optimization &
IPSM with depth/geometric regularization &
Compatible LLFF and DTU settings &
Public reported values marked \\
\hline
Tang et al.~\cite{tang2024finetuning} &
NeRF-based sparse-view reconstruction &
Fine-tuned diffusion prior with coarse rendering/text prompts &
No strictly aligned 3DGS benchmark protocol identified &
Discussed in Related Work only \\
\hline
\end{tabular}
\vspace{-2mm}
\end{table*}

\begin{table*}[t]
\centering
\caption{Quantitative comparison on LLFF under 3-view, 6-view, and 9-view settings.}
\label{tab:r1_llff_comparison}
\scriptsize
\setlength{\tabcolsep}{2.5pt}
\resizebox{1.0\textwidth}{!}{%
\begin{tabular}{llccccccccc}
\hline
\textbf{Method} & \textbf{Result source} &
\multicolumn{3}{c}{\textbf{3-view}} &
\multicolumn{3}{c}{\textbf{6-view}} &
\multicolumn{3}{c}{\textbf{9-view}} \\
\cline{3-5}\cline{6-8}\cline{9-11}
 & &
\textbf{LPIPS* \(\downarrow\)} & \textbf{PSNR \(\uparrow\)} & \textbf{SSIM* \(\uparrow\)} &
\textbf{LPIPS* \(\downarrow\)} & \textbf{PSNR \(\uparrow\)} & \textbf{SSIM* \(\uparrow\)} &
\textbf{LPIPS* \(\downarrow\)} & \textbf{PSNR \(\uparrow\)} & \textbf{SSIM* \(\uparrow\)} \\
\hline
IPSM-Gaussian~\cite{wang2024diffusionpriors} & Reported & 20.70 & 20.44 & 70.20 & 13.50 & 23.94 & 81.80 & 11.10 & 25.13 & 85.50 \\
CoR-GS~\cite{zhang2024corgs} & Reported & 19.60 & 20.45 & 71.20 & 11.50 & 24.49 & 83.70 & 8.90 & 26.06 & 87.40 \\
NexusGS~\cite{zheng2025nexusgs} & Reported / Re-run & 17.70 & 21.07 & 73.80 & 10.65 & 24.83 & 84.92 & 8.35 & 26.24 & 88.00 \\
\textbf{HierGF (ours)} & \textbf{Ours} & \textbf{17.05} & \textbf{21.28} & \textbf{74.45} & \textbf{10.10} & \textbf{25.03} & \textbf{85.42} & \textbf{7.97} & \textbf{26.39} & \textbf{88.37} \\
\hline
\end{tabular}%
}
\vspace{-2mm}
\end{table*}

\begin{table*}[t]
\centering
\caption{Quantitative comparison on DTU under 3-view, 6-view, and 9-view settings.}
\label{tab:r1_dtu_comparison}
\scriptsize
\setlength{\tabcolsep}{2.5pt}
\resizebox{\textwidth}{!}{%
\begin{tabular}{llccccccccc}
\hline
\textbf{Method} & \textbf{Result source} &
\multicolumn{3}{c}{\textbf{3-view}} &
\multicolumn{3}{c}{\textbf{6-view}} &
\multicolumn{3}{c}{\textbf{9-view}} \\
\cline{3-5}\cline{6-8}\cline{9-11}
 & &
\textbf{LPIPS* \(\downarrow\)} & \textbf{PSNR \(\uparrow\)} & \textbf{SSIM* \(\uparrow\)} &
\textbf{LPIPS* \(\downarrow\)} & \textbf{PSNR \(\uparrow\)} & \textbf{SSIM* \(\uparrow\)} &
\textbf{LPIPS* \(\downarrow\)} & \textbf{PSNR \(\uparrow\)} & \textbf{SSIM* \(\uparrow\)} \\
\hline
IPSM-Gaussian~\cite{wang2024diffusionpriors} & Reported & 12.10 &19.99 & 85.60 & -- & -- & -- & -- & -- & -- \\
CoR-GS~\cite{zhang2024corgs} & Reported & 11.90 & 19.21 & 85.30 & 6.80 & 24.51 & 91.70 & 4.50 & 27.18 & 94.70 \\
NexusGS~\cite{zheng2025nexusgs} & Reported / Re-run & 10.20 & 20.21 & 86.90 & 6.15 & 24.91 & 92.25 & 4.15 & 27.38 & 95.05 \\
\textbf{HierGF (ours)} & \textbf{Ours} & \textbf{9.68} & \textbf{20.40} & \textbf{87.45} & \textbf{5.74} & \textbf{25.10} & \textbf{92.68} & \textbf{3.89} & \textbf{27.52} & \textbf{95.36} \\
\hline
\end{tabular}%
}
\vspace{-3mm}
\end{table*}

Additional LLFF and DTU comparisons are reported in Tables~\ref{tab:r1_llff_comparison} and~\ref{tab:r1_dtu_comparison}, respectively, to complement the main MipNeRF360, OmniObject3D, and OpenIllumination evaluations with commonly used sparse-view benchmarks.

\begin{figure*}[htbp]
  \centering
  \IfFileExists{second_round_review_response/figures/LLFF.pdf}{%
      \includegraphics[width=0.85\textwidth]{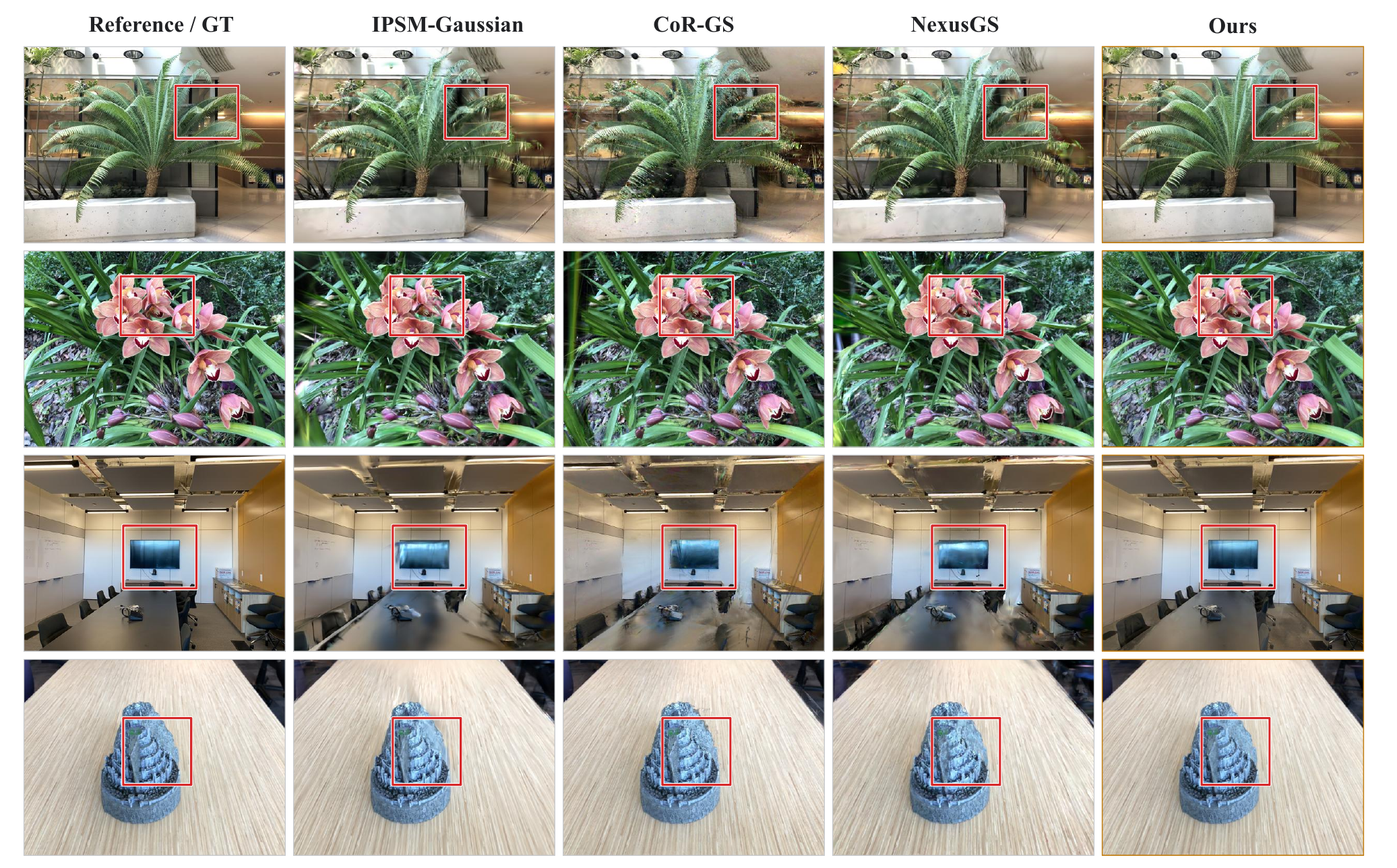}%
  }{%
      \rule{0pt}{0.29\textwidth}%
  }
  \caption{Qualitative comparison on LLFF under the 3-view setting. Representative held-out views from \texttt{fern}, \texttt{orchids}, \texttt{room}, and \texttt{fortress} are shown. Columns compare the ground truth, IPSM-Gaussian, CoR-GS, NexusGS, and HierGF at aligned viewpoints; enlarged crops emphasize thin structures, boundaries, texture details, and weakly observed regions.}
  \label{fig:llff_qualitative}
\end{figure*}

\begin{figure*}[htbp]
  \centering
  \IfFileExists{second_round_review_response/figures/DTU.pdf}{%
      \includegraphics[width=0.85\textwidth]{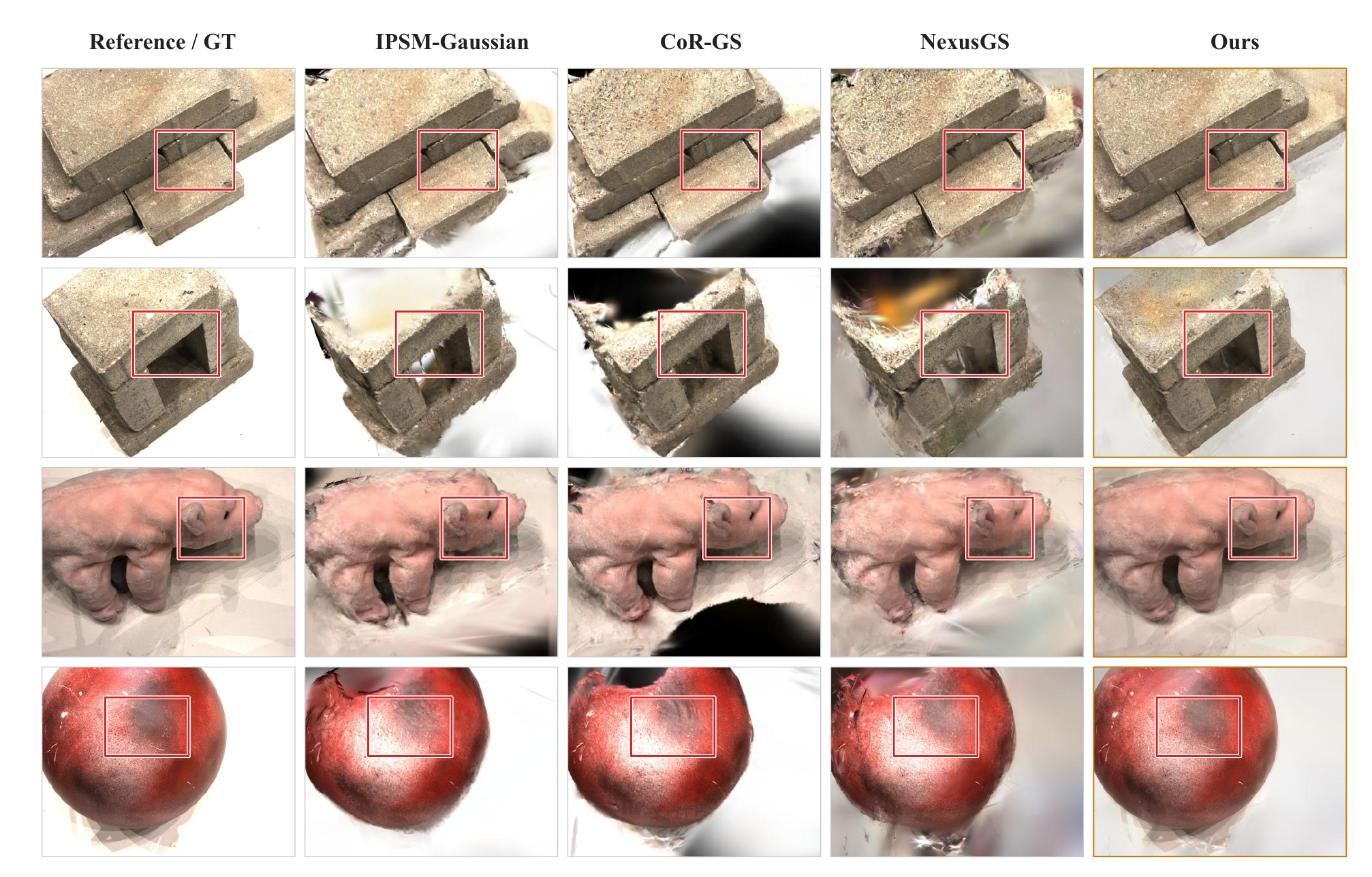}%
  }{%
      \rule{0pt}{0.29\textwidth}%
  }
  \caption{Qualitative comparison on DTU under the 3-view setting. Representative held-out views from scans 34, 40, 103, and 8 are shown. Columns compare the ground truth, IPSM-Gaussian, CoR-GS, NexusGS, and HierGF at aligned viewpoints; enlarged crops emphasize object contours, fine geometry, surface texture, and sparsely observed regions.}
  \label{fig:dtu_qualitative}
\end{figure*}

To complement these quantitative results, Figs.~\ref{fig:llff_qualitative} and~\ref{fig:dtu_qualitative} provide aligned qualitative comparisons under the challenging 3-view setting. The LLFF examples cover the \texttt{fern}, \texttt{orchids}, \texttt{room}, and \texttt{fortress} scenes, while the DTU examples cover scans 34, 40, 103, and 8. We compare CoR-GS, NexusGS, IPSM-Gaussian, and HierGF against the corresponding ground truth and use consistent enlarged crops to facilitate inspection of thin structures, boundaries, texture details, and weakly observed regions.

Across MipNeRF360, OmniObject3D and OpenIllumination, HierGF achieves the best or very competitive PSNR and SSIM under all tested view counts, and on the challenging 4-view setting it often improves PSNR by more than ten percent over the strongest baselines, showing that the hierarchical geometry–perception design can still recover high-fidelity geometry and appearance when multi-view constraints are weak. In terms of LPIPS, HierGF surpasses the strongest baselines on most scenes, especially for four and six views where it leads on the majority of cases and stays within a modest gap in the remaining ones, which aligns with our goal of turning coarse geometry and 2D priors into reliable pseudo-supervision rather than only matching pixels on sparse reference views.

Qualitative comparisons in Fig.~\ref{fig:vis1} and Fig.~\ref{fig:open} further support these findings. On unbounded indoor and outdoor scenes from MipNeRF360, HierGF produces sharper distant structures, cleaner backgrounds and fewer floaters than NeRF-based baselines and vanilla 3DGS. On the diverse objects in OmniObject3D, our method preserves fine details such as thin handles, leaves and legs that are often over-smoothed or fragmented by other methods under sparse supervision. On OpenIllumination, which exhibits strong lighting variation, our reconstructions show more consistent shading and fewer intensity jumps across views, suggesting that the 2D-assisted perceptual supervision and confidence weighting help to stabilize appearance in under-constrained regions.

\begin{figure}
\vspace{-3mm}
\centering
\includegraphics[width=0.8\columnwidth]{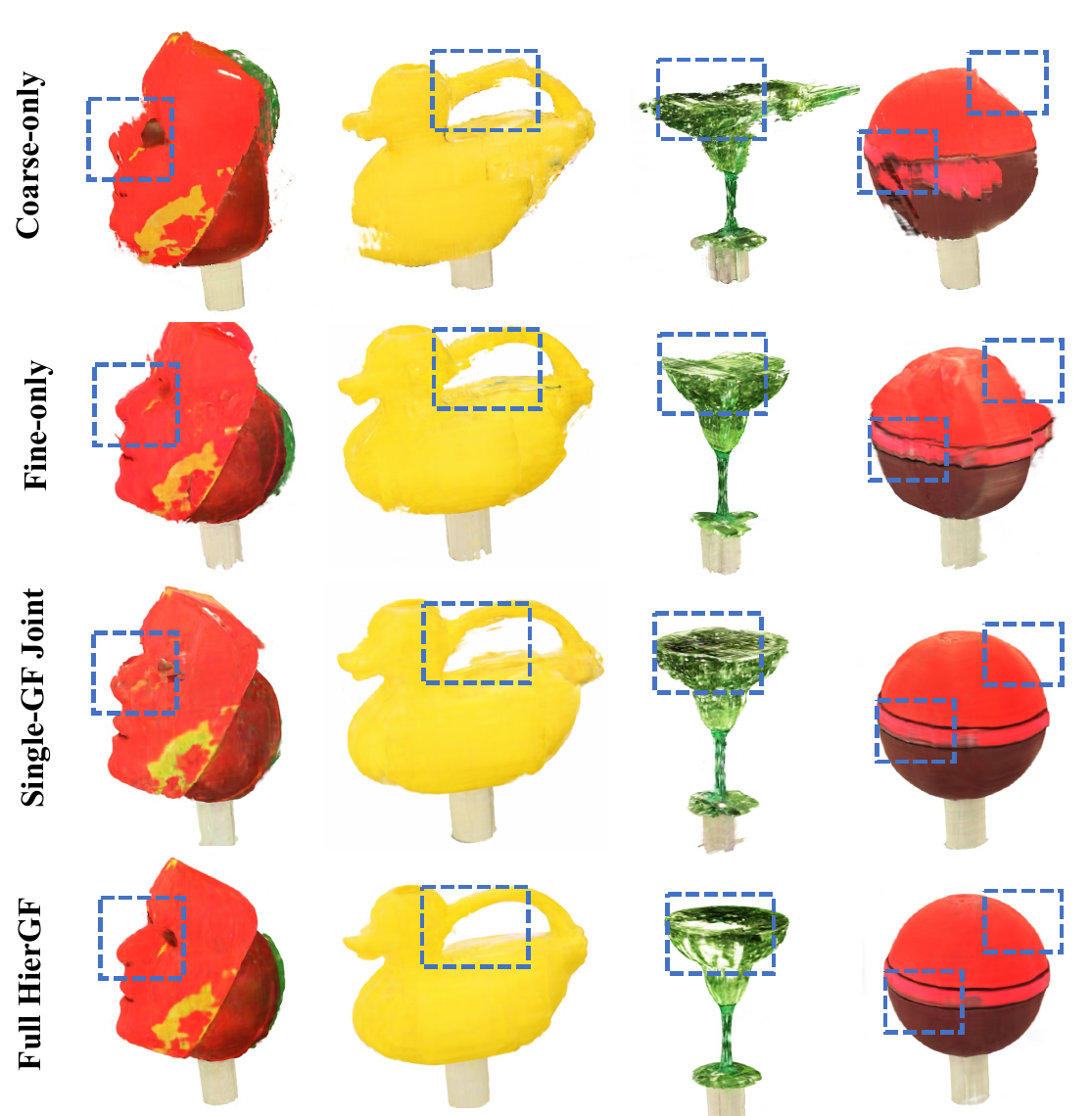}
\caption{Ablation study on the two-stage hierarchical backbone. From top to bottom, Coarse-only, Fine-only, Single-GF Joint, and Full HierGF are shown.}
\label{fig:ablation_hie}
\end{figure}

\begin{figure}
\centering
\vspace{-2mm}
\includegraphics[width=0.85\columnwidth]{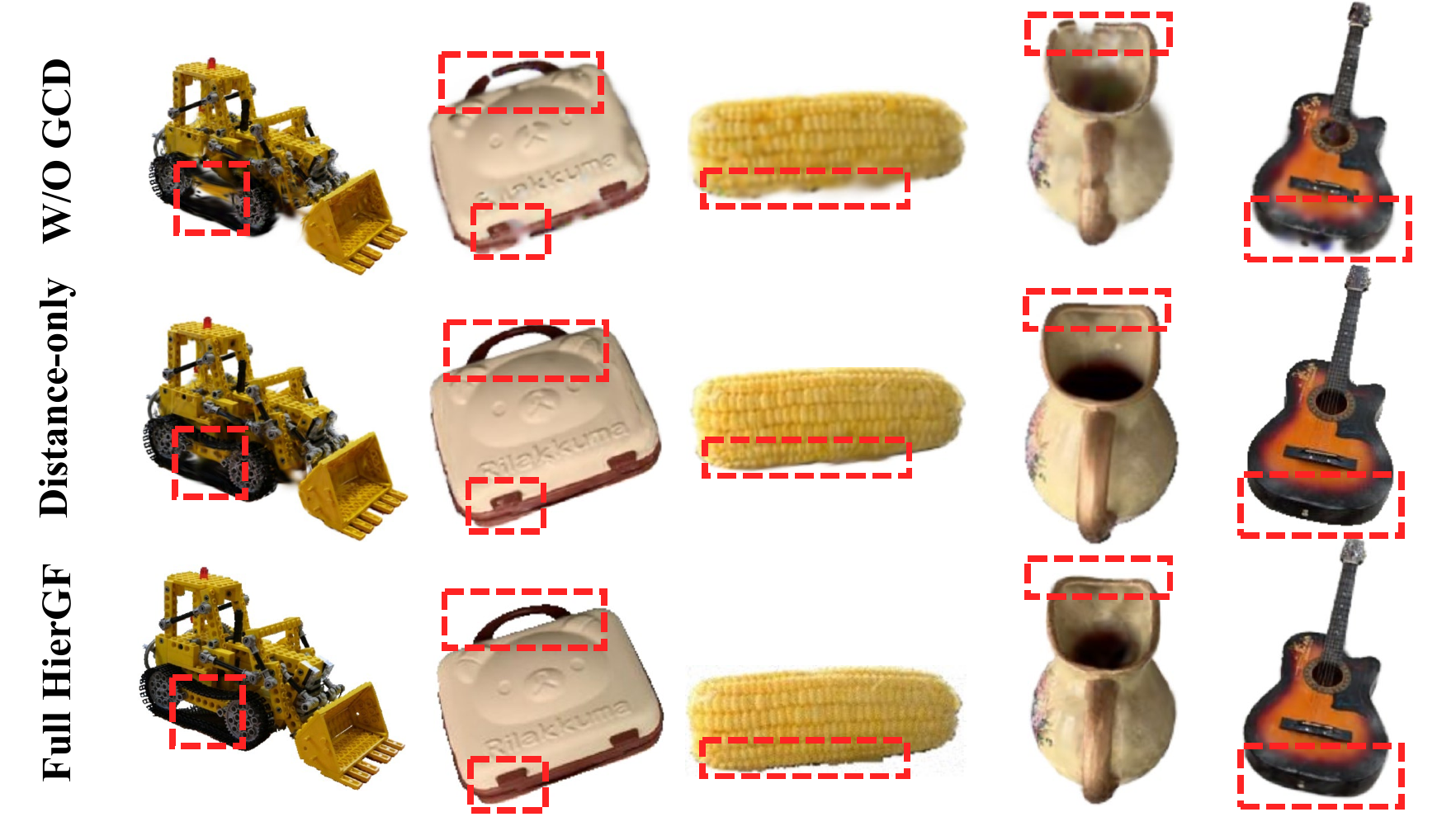}
\caption{Ablation study on geometry-consistent densification. Red boxes highlight that our full model better preserves thin structures and closes gaps in under-sampled regions.}
\vspace{-2mm}
\label{fig:ablation_gcd}
\vspace{-4mm}
\end{figure}

\begin{table}
\caption{Ablation study over HierGF's hierarchical architecture on the MipNeRF360 and OmniObject datasets with 4 input views.}
\label{tab:ablation_hie}
\centering
\resizebox{\columnwidth}{!}{
\begin{tabular}{@{}c@{}c@{}c|ccc|ccc@{}}
\toprule
& & \multicolumn{1}{c|}{ \multirow{2.5}{*}{\textbf{Setting/Dataset}}}  & \multicolumn{3}{c|}{\textbf{MipNeRF360}} & \multicolumn{3}{c}{\textbf{OmniObject}} \\ 
\cmidrule(r){4-6} \cmidrule(l){7-9} 
&  &  & LPIPS$^{*}$ ↓ & PSNR ↑ & SSIM$^{*}$ ↑ & LPIPS$^{*}$ ↓ & PSNR ↑ & SSIM$^{*}$ ↑ \\
\midrule

\multirow{10}{*} &  & Coarse-only & 9.76 & 19.63 & 88.90 & 7.92 & 17.88 & 94.35\\
\midrule

\multirow{10}{*} &  & Fine-only & 8.12 & 21.84 & 91.03 & 6.84 & 20.59 & 95.01\\
\midrule

\multirow{10}{*} &  & Single-GF Joint & 5.24 & 25.06 & 92.95 & 3.43 & 27.06 & 96.90\\
\midrule

\multirow{10}{*} &  & Full HierGF  & \textbf{3.85} & \textbf{28.48} & \textbf{95.87} & \textbf{1.76} & \textbf{34.08} & \textbf{98.36}\\


\bottomrule
\end{tabular}}
\vspace{-3mm}

\end{table}

\subsection{Ablation Studies}
We conduct ablation experiments under identical data and training settings to analyze how the two-stage hierarchical backbone, selective coarse-to-fine transfer, ZoeDepth prior, pseudo-view enhancement, diffusion prior, LoRA adaptation, confidence weighting, and geometry-consistent densification jointly influence sparse-view reconstruction performance. Unless otherwise stated, controlled comparisons keep the input views, camera poses, losses, pseudo-view refresh schedule, GCD configuration, and optimization budget unchanged while varying only the examined factor.

\noindent \textbf{Ablation Study on the two-stage hierarchical backbone.} 
As shown in Tab.~\ref{tab:ablation_hie} and Fig.~\ref{fig:ablation_hie}, we study the two-stage design of HierGF under identical cameras, losses and total iterations. In the coarse only variant, a single Gaussian field is trained on sparse views with color and monocular depth losses and standard adaptive density control, without pseudo-views, diffusion enhancement, confidence weighting or geometry-consistent density completion, which behaves like a 3DGS-style backbone and recovers the global shape but keeps under-sampled regions blurry and thin structures broken. In the fine-only variant, we remove the separate coarse stage and run the second stage procedure from the initial field, so early pseudo views are often misaligned and the diffusion prior may inject plausible but wrong details, giving sharper local textures but unstable geometry and lower PSNR and SSIM than HierGF. We also test a single field joint training variant that uses one Gaussian field for the whole optimization and applies 2D-assisted perceptual supervision every $K$ iterations with $K = 500$, which reduces some artifacts compared with fine-only but still falls behind HierGF in perceptual quality and multi-view consistency, showing that both stages and the explicit hierarchy are needed.
To isolate the transfer rule within the two-field backbone, Table~\ref{tab:r1_c1_selective_transfer} compares different fine-field initialization strategies while keeping the two-stage schedule, pseudo-view supervision, confidence weighting, GCD setting, and optimization budget unchanged.

\begin{table}
\caption{Ablation study over 2D-assisted Perceptual Enhancement on the MipNeRF360 and dataset with 4 input views. \textit{SPS}, \textit{VCS}, and \textit{PCS} represent self-generated perceptual supervision, view-level confidence supervision and pixel-level confidence supervision, respectively.}
\label{tab:ablation2}
\centering
\resizebox{0.7\columnwidth}{!}{
\begin{tabular}{@{}c@{}c@{}ccc|ccc@{}}
\toprule
& & \multicolumn{3}{c|}{ {\textbf{Components}}}  & \multicolumn{3}{c}{\textbf{Metrics}} \\ 
\cmidrule(r){1-5}  \cmidrule(l){6-8}
& & SPS & VCS & PCS & LPIPS$^{*}$ $\downarrow$ & PSNR $\uparrow$ & SSIM$^{*}$ $\uparrow$\\
\midrule

\multirow{10}{*} &  &  &  &  & 9.40 & 20.33 & 89.78 \\
\midrule
\multirow{10}{*} &  & \checkmark &  &  & 5.06 & 25.73 & 92.18 \\
\midrule
\multirow{1}{*} &  & \checkmark & \checkmark &  & 4.71 & 27.62 & 93.04 \\
\midrule
\multirow{1}{*} &  & \checkmark & & \checkmark & \textbf{3.13} & 28.09 & 95.26 \\
\midrule
\multirow{10}{*} &  & \checkmark & \checkmark & 
\checkmark & 3.85 & \textbf{28.48} & \textbf{95.87} \\

\bottomrule
\end{tabular}}
\vspace{-2mm}
\end{table}

\begin{table}
\caption{Ablation study of geometry-consistent densification on MipNeRF360 and OmniObject3D with four input views. \textit{GCD} denotes geometry-consistent densification}.
\label{tab:ablation_gcd}
\centering
\resizebox{\columnwidth}{!}{
\begin{tabular}{@{}c@{}c@{}c|ccc|ccc@{}}
\toprule
& & \multicolumn{1}{c|}{ \multirow{2.5}{*}{\textbf{Setting/Dataset}}}  & \multicolumn{3}{c|}{\textbf{MipNeRF360}} & \multicolumn{3}{c}{\textbf{OmniObject}} \\ 
\cmidrule(r){4-6} \cmidrule(l){7-9} 
&  &  & LPIPS$^{*}$ ↓ & PSNR ↑ & SSIM$^{*}$ ↑ & LPIPS$^{*}$ ↓ & PSNR ↑ & SSIM$^{*}$ ↑ \\
\midrule

\multirow{10}{*} &  & w/o GCD & 4.58 & 27.96 & 93.37 & 1.93 & 31.07 & 97.56 \\
\midrule

\multirow{10}{*} &  & Distance-only & 4.02 & 28.25 & 94.92 & 1.79 & 33.74 & 98.05\\
\midrule

\multirow{10}{*} &  & Full HierGF  & \textbf{3.85} & \textbf{28.48} & \textbf{95.87} & \textbf{1.76} & \textbf{34.08} & \textbf{98.36}\\


\bottomrule
\end{tabular}}
\vspace{-5mm}
\end{table}

\begin{table*}[t]
\centering
\caption{Controlled ablation of the coarse-to-fine parameter-transfer rule under the 4-view setting.}
\label{tab:r1_c1_selective_transfer}
\scriptsize
\setlength{\tabcolsep}{3pt}
\renewcommand{\arraystretch}{1.18}
\resizebox{\textwidth}{!}{%
\begin{tabular}{p{3.8cm}p{4.4cm}cccccc}
\hline
\multirow{2}{*}{\parbox{3.8cm}{\centering\textbf{Fine-field initialization}\\\textbf{strategy}}} &
\multirow{2}{*}{\parbox{4.4cm}{\centering\textbf{Transferred parameters}\\\textbf{from coarse field}}} &
\multicolumn{3}{c}{\textbf{MipNeRF360}} &
\multicolumn{3}{c}{\textbf{OmniObject3D}} \\
\cline{3-8}
& &
\textbf{LPIPS* \(\downarrow\)} & \textbf{PSNR \(\uparrow\)} & \textbf{SSIM* \(\uparrow\)} &
\textbf{LPIPS* \(\downarrow\)} & \textbf{PSNR \(\uparrow\)} & \textbf{SSIM* \(\uparrow\)} \\
\hline
Reinitialize all attributes &
None &
4.46 & 27.21 & 94.67 &
2.21 & 31.65 & 97.53 \\
\hline
Transfer all attributes &
\(\mu, q, s, \sigma, \mathrm{SH}\) &
4.18 & 27.82 & 95.14 &
1.98 & 32.93 & 97.95 \\
\hline
\textbf{Selective transfer, HierGF} &
\textbf{\(\mu\) and RGB only} &
\textbf{3.85} & \textbf{28.48} & \textbf{95.87} &
\textbf{1.76} & \textbf{34.08} & \textbf{98.36} \\
\hline
\end{tabular}%
}
\vspace{-2mm}
\end{table*}

\begin{table*}[t]
\centering
\caption{Controlled ablation of ZoeDepth supervision under the 4-view setting.}
\label{tab:r2_c2_zoedepth}
\small
\setlength{\tabcolsep}{5.5pt}
\setlength{\extrarowheight}{1pt}
\renewcommand{\arraystretch}{1.15}
\begin{tabular}{@{}lcccccc@{}}
\hline
\textbf{Setting} &
\multicolumn{3}{c}{\textbf{MipNeRF360}} &
\multicolumn{3}{c}{\textbf{OmniObject3D}} \\
\cline{2-4}\cline{5-7}
&
\textbf{LPIPS$^{*}\downarrow$} &
\textbf{PSNR$\uparrow$} &
\textbf{SSIM$^{*}\uparrow$} &
\textbf{LPIPS$^{*}\downarrow$} &
\textbf{PSNR$\uparrow$} &
\textbf{SSIM$^{*}\uparrow$} \\
\hline
Full HierGF w/o ZoeDepth &
4.29 & 27.72 & 95.01 &
1.91 & 33.43 & 98.02 \\
\textbf{Full HierGF} &
\textbf{3.85} & \textbf{28.48} & \textbf{95.87} &
\textbf{1.76} & \textbf{34.08} & \textbf{98.36} \\
\hline
\end{tabular}
\vspace{-2mm}
\end{table*}

\begin{table*}[t]
\centering
\caption{Controlled ablation of pseudo-view supervision, the diffusion prior, LoRA adaptation, and confidence weighting on MipNeRF360 under the 4-view setting.}
\label{tab:r2_c2_perceptual}
\scriptsize
\setlength{\tabcolsep}{4pt}
\renewcommand{\arraystretch}{1.15}
\resizebox{0.8\textwidth}{!}{%
\begin{tabular}{lccc}
\hline
\textbf{Setting} & \textbf{LPIPS$^{*}\downarrow$} & \textbf{PSNR$\uparrow$} & \textbf{SSIM$^{*}\uparrow$} \\
\hline
Reference views only; no pseudo-view supervision & 10.07 & 19.86 & 89.05 \\
Raw pseudo-views; no diffusion enhancement or confidence weighting & 9.40 & 20.33 & 89.78 \\
Frozen pre-trained 2D prior; without LoRA adaptation or confidence weighting & 5.73 & 24.86 & 91.64 \\
LoRA-adapted enhanced pseudo-views (SPS); without confidence weighting & 5.06 & 25.73 & 92.18 \\
SPS + view-level confidence supervision (VCS) & 4.71 & 27.62 & 93.04 \\
SPS + pixel-level confidence supervision (PCS) & \textbf{3.13} & 28.09 & 95.26 \\
\textbf{SPS + VCS + PCS (Full HierGF)} & 3.85 & \textbf{28.48} & \textbf{95.87} \\
\hline
\end{tabular}%
}
\vspace{-2mm}
\end{table*}

\begin{table*}[t]
\centering
\caption{Ablation study on the neighborhood radius \(r\) under the 4-view setting.}
\label{tab:r1_radius_ablation}
\scriptsize
\setlength{\tabcolsep}{3pt}
\resizebox{0.8\textwidth}{!}{%
\begin{tabular}{lccccccc}
\hline
\textbf{Radius setting} & \textbf{Radius value} &
\multicolumn{3}{c}{\textbf{MipNeRF360}} &
\multicolumn{3}{c}{\textbf{OmniObject3D}} \\
\cline{3-5}\cline{6-8}
 & &
\textbf{LPIPS* \(\downarrow\)} & \textbf{PSNR \(\uparrow\)} & \textbf{SSIM* \(\uparrow\)} &
\textbf{LPIPS* \(\downarrow\)} & \textbf{PSNR \(\uparrow\)} & \textbf{SSIM* \(\uparrow\)} \\
\hline
Small radius & \(1.5s_v=0.03D_{\mathrm{rob}}\) & 4.16 & 28.17 & 94.82 & 1.86 & 
33.72 & 98.04 \\
Default radius & \(2.0s_v=0.04D_{\mathrm{rob}}\) & \textbf{3.85} & \textbf{28.48} & \textbf{95.87} & \textbf{1.76} & \textbf{34.08} & \textbf{98.36} \\
Large radius & \(3.0s_v=0.06D_{\mathrm{rob}}\) & 4.04 & 28.28 & 95.18 & 1.83 & 33.89 & 98.18 \\
\hline
\end{tabular}%
}
\vspace{-4mm}
\end{table*}

\begin{table*}[t]
\centering
\vspace{-4mm}
\caption{Comparison of weighting strategies in geometry-consistent densification under the 4-view setting.}
\label{tab:r1_weighting_ablation}
\scriptsize
\setlength{\tabcolsep}{3pt}
\resizebox{0.75\textwidth}{!}{%
\begin{tabular}{lcccccc}
\hline
\textbf{Setting} &
\multicolumn{3}{c}{\textbf{MipNeRF360}} &
\multicolumn{3}{c}{\textbf{OmniObject3D}} \\
\cline{2-4}\cline{5-7}
 &
\textbf{LPIPS* \(\downarrow\)} & \textbf{PSNR \(\uparrow\)} & \textbf{SSIM* \(\uparrow\)} &
\textbf{LPIPS* \(\downarrow\)} & \textbf{PSNR \(\uparrow\)} & \textbf{SSIM* \(\uparrow\)} \\
\hline
w/o GCD, standard ADC & 4.58 & 27.96 & 93.37 & 1.93 & 31.07 & 97.56 \\
Distance-only interpolation & 4.02 & 28.25 & 94.92 & 1.79 & 33.74 & 98.05 \\
Additive: \(w_{\mathrm{normal}}+w_{\mathrm{distance}}\) & 3.95 & 28.34 & 95.31 & 1.78 & 
33.91 & 98.18 \\
Multiplicative: \(w_{\mathrm{normal}}\cdot w_{\mathrm{distance}}\) & \textbf{3.85} & \textbf{28.48} & \textbf{95.87} & \textbf{1.76} & \textbf{34.08} & \textbf{98.36} \\
\hline
\end{tabular}%
}
\vspace{-2mm}
\end{table*}

\noindent \textbf{Ablation on 2D-assisted perceptual enhancement.} 
We further assess the effect of 2D-assisted perceptual enhancement under the full two-stage HierGF backbone, as reported in Tab.~\ref{tab:ablation2}. When 2D assistance is removed, pseudo-views are still rendered from the coarse Gaussian field every $K=500$ iterations and used as extra supervision, but direct training on raw images leaves occluded and low-light regions unrealistic and further degrades cross-view shading. Dropping pixel-level confidence or view-level confidence also reduces LPIPS and cross-view SSIM, since unreliable pixels and globally inconsistent pseudo-views cannot be suppressed, whereas the full 2D-assisted setting, which combines diffusion-based enhancement with both confidence terms, achieves the best perceptual quality and multi-view consistency on all three datasets.

Table~\ref{tab:r2_c2_zoedepth} isolates the contribution of ZoeDepth supervision under the full HierGF configuration, showing that the monocular depth prior stabilizes sparse-view geometry but is not the sole source of the hierarchical refinement gain.

Table~\ref{tab:r2_c2_perceptual} further separates the effects of additional pseudo-views, the frozen diffusion prior, scene-specific LoRA adaptation, and pixel-/view-level confidence weighting, showing that enhanced pseudo-supervision is most effective when combined with confidence-guided filtering.

\noindent \textbf{Ablation Study on geometry-consistent densification.}
We further evaluate the geometry-consistent density completion under the full two-stage HierGF with 2D-assisted perceptual supervision. When this module is removed, that is, we disable voxel-based sparse-region detection and normal-weighted interpolation and rely only on standard 3DGS density control, the global shape is recovered but under-sampled areas exhibit gaps, floaters, and broken thin structures, with clear drops in SSIM and LPIPS in Tab.~\ref{tab:ablation_gcd}. A Distance-only variant, which still detects sparse voxels and inserts new Gaussians but uses distance-based interpolation without normal similarity, reduces some holes but produces point drift across surface layers and softened edges, as shown in Fig.~\ref{fig:ablation_gcd}. The full geometry-consistent density completion, combining sparse-voxel detection with distance and normal-aware interpolation, yields the most stable geometry and best preserves fine details under the same training budget.

The radius sensitivity of geometry-consistent densification is reported in Table~\ref{tab:r1_radius_ablation}, where the default radius balances stable PCA-based normal estimation with local-structure preservation.

Table~\ref{tab:r1_weighting_ablation} compares distance-only, additive normal-distance, and multiplicative normal-distance weighting, validating the multiplicative formulation used by HierGF for structure-aware Gaussian insertion.

\subsection{Efficiency Analysis}
\begin{figure*}[t]
\centering
\includegraphics[width=0.9\textwidth]{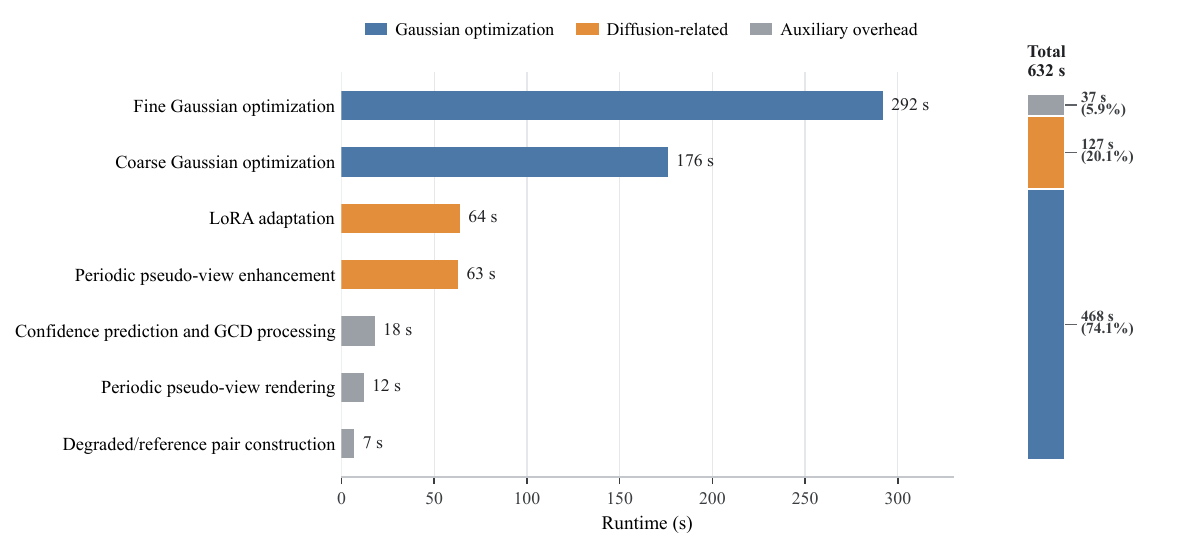}
\vspace{-2mm}
\caption{Stage-wise runtime breakdown of HierGF on the MipNeRF360 \texttt{kitchen} scene under the 4-view setting. Gaussian optimization remains the dominant cost, while diffusion-related processing is used only during reconstruction.}
\label{fig:r2_c4_runtime_bar}
\vspace{-2mm}
\end{figure*}

Table~\ref{tab:r2_c4_efficiency} reports per-scene reconstruction time and post-optimization novel-view rendering time on the MipNeRF360 \texttt{kitchen} scene under the 4-view setting. All runtimes are measured at a resolution of \(779\times520\) on a single NVIDIA A100 GPU. HierGF requires 632~s for reconstruction and 70~s for rendering the held-out evaluation views after optimization. Relative to ZeroRF, GaussianObject, and WaveletGS, HierGF achieves reconstruction speedups of \(2.07\times\), \(11.12\times\), and \(2.85\times\), respectively. Its post-optimization rendering time is reduced by \(10.3\%\), \(21.3\%\), and \(26.3\%\) relative to these methods. The stage-wise breakdown in Fig.~\ref{fig:r2_c4_runtime_bar} further shows that the diffusion module is used only during reconstruction for pseudo-view enhancement; final rendering relies solely on the optimized Gaussian representation.

\begin{table*}[t]
\centering
\caption{Efficiency comparison on the MipNeRF360 \texttt{kitchen} scene under the 4-view setting. All methods are evaluated at \(779\times520\) on a single NVIDIA A100 GPU. The speedup values are calculated relative to HierGF.}
\label{tab:r2_c4_efficiency}
\small
\setlength{\tabcolsep}{5pt}
\setlength{\extrarowheight}{1pt}
\renewcommand{\arraystretch}{1.18}
\begin{tabular}{@{}
>{\raggedright\arraybackslash}p{0.22\textwidth}
c
c
c
c
@{}}
\hline
\textbf{Method} &
\textbf{Recon. time (s) \(\downarrow\)} &
\textbf{Recon. speedup} &
\textbf{Rendering time (s) \(\downarrow\)} &
\textbf{Rendering speedup} \\
\hline
ZeroRF~\cite{shi2024zerorf} &
1309 &
\(2.07\times\) &
78 &
\(1.11\times\) \\
GaussianObject~\cite{yang2024gaussianobject} &
7028 &
\(11.12\times\) &
89 &
\(1.27\times\) \\
WaveletGS~\cite{nguyen2025waveletgaussian} &
1800 &
\(2.85\times\) &
95 &
\(1.36\times\) \\
\textbf{HierGF (ours)} &
\textbf{632} &
-- &
\textbf{70} &
-- \\
\hline
\end{tabular}
\vspace{-4mm}
\end{table*}

%
\vspace{-0.1in}
\section{Conclusion}



In this paper, we present HierGF for sparse-view 3D reconstruction, trained with a two-stage geometry-perception paradigm that renders pseudo-views from a coarse Gaussian field, refines them with a frozen-backbone, LoRA-adapted diffusion prior and confidence weighting, and uses normal-weighted densification to preserve surface continuity in under-sampled regions. On three primary sparse-view benchmarks, with additional quantitative comparisons on LLFF and DTU, HierGF improves reconstruction quality and cross-view consistency from few inputs. The revised efficiency analysis further shows that HierGF completes per-scene reconstruction in 632~s and post-optimization rendering in 70~s on the MipNeRF360 \texttt{kitchen} 4-view setting using a single NVIDIA A100 GPU. In future work, we will consider replacing interpolation with learnable geometric priors and extending the method to unposed or dynamic scenes.
\vspace{-2mm}

\bibliographystyle{IEEEtran}
\bibliography{ref}

\newpage

 




\vfill

\end{document}